\documentclass{article} 
\usepackage{iclr2022_conference,times}

\usepackage{amsmath,amsfonts,bm}

\def\eqref#1{equation~\ref{#1}}

\def\1{\bm{1}}

\DeclareMathAlphabet{\mathsfit}{\encodingdefault}{\sfdefault}{m}{sl}
\SetMathAlphabet{\mathsfit}{bold}{\encodingdefault}{\sfdefault}{bx}{n}

\DeclareMathOperator*{\argmin}{arg\,min}

\usepackage{hyperref}
\usepackage{url}
\usepackage{amssymb}
\newcommand{\methodname}{Locality Sensitive Pruning}
\usepackage{pgfplots}
\pgfplotsset{compat=1.17}
\usetikzlibrary{pgfplots.groupplots}
\usepackage{caption}
\usepackage{subcaption}
\usepackage{algorithm}
\usepackage{algpseudocode}
\usepackage{tabularx}
\usepackage{graphicx}
\usepackage{makecell}
\usepackage{wrapfig, blindtext}

\def\minhash{\emph{MinHash}}
\def\Real{\mathbb{R}}
\def\Natural{\mathbb{N}}
\def\neighborhood{\mathcal{N}}

\def\mathH{\mathcal{H}}

\newcommand{\qmheight}{1.15in}

\newcommand{\timelabel}{Normalized Time}

\newtheorem{thm}{Theorem}
\newtheorem{example}[thm]{Example}

\newtheorem{proposition}[thm]{Proposition}
\newtheorem{definition}[thm]{Definition}

\newcommand{\myxticks}{{0,0.2,0.4,0.6,0.8,1}}

\def\notes{} 
\ifdefined\notes
\newcommand{\note}[2]{{\color{#1}{#2}}}
\newcommand\jnote[1]{ \note{purple}{[Joel: #1]}}
\newcommand\enote[1]{ \note{red}{[Eitan: #1]}}
\newcommand\dnote[1]{ \note{blue}{[Dotan: #1]}}
\else

\newcommand{\note}[2]{}
\newcommand\jnote[1]{ }
\newcommand\enote[1]{ }
\newcommand\dnote[1]{ }
\fi

\title{LSP : Acceleration and Regularization of Graph Neural Networks via Locality Sensitive Pruning of Graphs}

\author{Eitan Kosman \\
Bosch Center for Artificial Intelligence\\
Haifa, Israel\\
\texttt{eitan.kosman@il.bosch.com} \\
\And
Joel Oren \\
Bosch Center for Artificial Intelligence \\
Haifa, Israel \\
\texttt{joel.oren@il.bosch.com} \\
\AND
Dotan Di Castro \\
Bosch Center for Artificial Intelligence \\
Haifa, Israel\\
\texttt{Dotan.DiCastro@il.bosch.com}
}

\iclrfinalcopy 
\begin{document}

\maketitle

\begin{abstract}\label{section:abstract}
Graph Neural Networks (GNNs) have emerged as highly successful tools for graph-related tasks. However, real-world problems involve very large graphs, and the compute resources needed to fit GNNs to those problems grow rapidly. Moreover, the noisy nature and size of real-world graphs cause GNNs to over-fit if not regularized properly. Surprisingly, recent works show that large graphs often involve many redundant components that can be removed without compromising the performance too much. This includes node or edge removals during inference through GNNs layers or as a pre-processing step that sparsifies the input graph. This intriguing phenomenon enables the development of state-of-the-art GNNs that are both efficient and accurate. In this paper, we take a further step towards demystifying this phenomenon and propose a systematic method called Locality-Sensitive Pruning (LSP) for graph pruning based on Locality-Sensitive Hashing. We aim to sparsify a graph so that similar local environments of the original graph result in similar environments in the resulting sparsified graph, which is an essential feature for graph-related tasks. To justify the application of pruning based on local graph properties, we exemplify the advantage of applying pruning based on locality properties over other pruning strategies in various scenarios. Extensive experiments on synthetic and real-world datasets demonstrate the superiority of LSP, which removes a significant amount of edges from large graphs without compromising the performance, accompanied by a considerable acceleration.
\end{abstract}

\section{Introduction}\label{section:introduction}
Graph neural networks have become extremely popular for tasks involving graph-data. The majority of architectures employ varieties of message-passings, such as Graph Convolutional Networks \citep{kipf2016semi}, Graph Isomorphism Networks \citep{xu2018powerful} and more \citep{wu2020comprehensive}. These architectures propagate belief information from nodes of a graph through adjacencies in order to generate representations that depend on wide graph environments. This property makes them well suited for learning tasks such as node classification \citep{bhagat2011node, zhang2018network}, link prediction \citep{kumar2020link} and graph classification \citep{kriege2020survey, cai2018comprehensive}. 

Although the aforementioned architectures have demonstrated great success, recent research has shown considerable limitations of GNNs. First, more complex architectures tend to be computationally demanding. For instance, Graph Attention Networks (GATs) \citep{velivckovic2017graph} whose neighborhood aggregation mechanism employ computations of self-attentions to assign weights for neighboring nodes. Second, this methodology of neighborhood aggregation leads to an exponentially growing amount of information originating from exponentially growing neighborhoods that needs to be encoded within fixed-size node representation vectors, a phenomenon referred as \textit{over-squashing} \citep{alon2020bottleneck}. Furthermore, the varying number of nodes participating in each node's neighborhood leads to a highly varying amount of information that needs to be encoded within a fixed-length code, a phenomenon that we call the \textit{varying neighborhoods}. Note that many other architectures (e.g., CNNs, MLPs, and RNNs) do not encounter this since they only accept fixed-size inputs or produce outputs with varying sizes that correspond to the input size. GNNs break this correspondence due to their neighborhood aggregation methodology. These phenomena become very prominent as the depth of the neighborhood grows, which limits our ability to develop deep GNNs. We further discuss these phenomena and provide explanations on how pruning edges could help mitigating them in Section \ref{section:general_advantagres}.

A popular approach for tackling these problems is to apply some transformation or augmentation to the input graph. One prominent approach is Graph Sparsification \citep{hu2013survey, spielman2008spectral, spielman2008graph, calandriello2018improved} in order to accelerate GNNs by removing nodes and edges from a graph under the constraint of preserving the predictive performance \citep{rong2019dropedge, srinivasa2020fast, ye2021sparse, hamilton2017inductive}. In fact, \cite{faber2021should} showed that very often a significant amount of edges can be removed without degrading the performance of a model, which raises a concern regarding the usage of popularly used benchmarks. They claimed that those tasks can be virtually solved through node features only, implying that the graph topology has limited contribution and edges can be safely disregarded.

\begin{figure}[!t]
    \centering
    \includegraphics[width=\linewidth, height=1.8in]{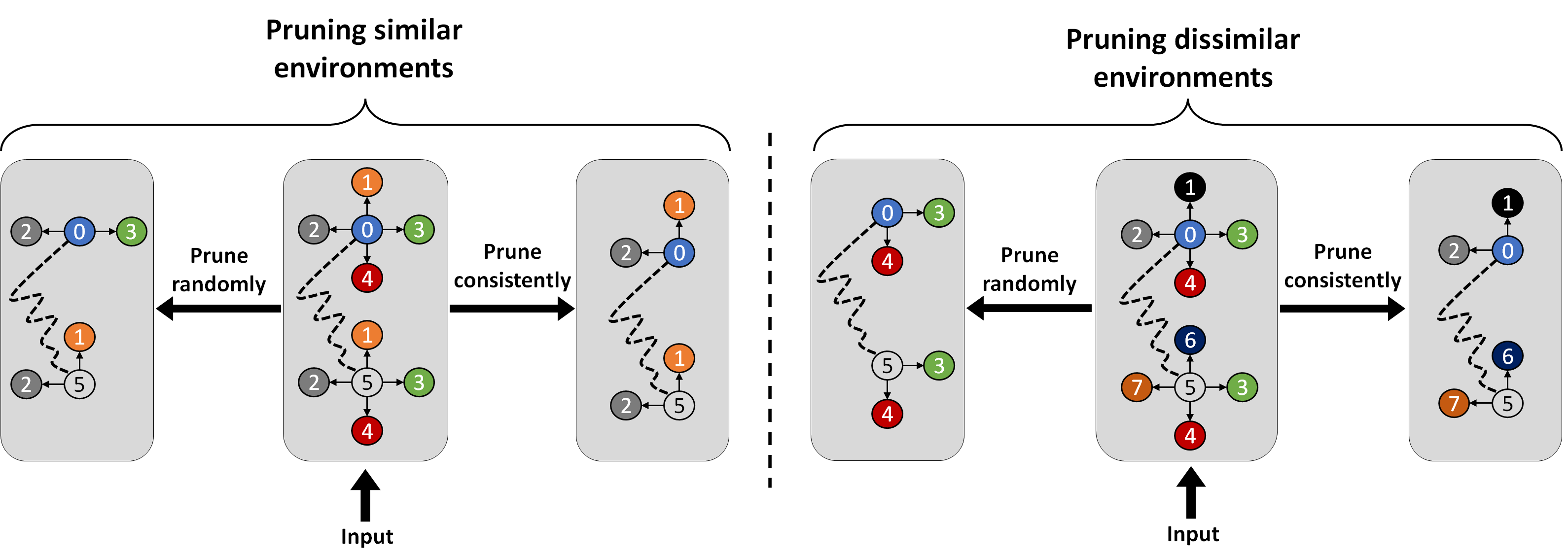}
    \caption{Illustration of consistent pruning versus non-consistent pruning. On the left, the input consists of a graph with two highlighted environments whose topology is similar because the central nodes have the same set of neighbors. In this case, consistent pruning results with a graph that preserves the similarity between these environments while non-consistent pruning, e.g., random removal of edges, results with dissimilar environments. On the right, the two highlighted environments have dissimilar topologies because the central nodes have different sets of neighbors. Consistent pruning is likely to preserve this dissimilarity, while \textit{random} might bring the similarity between them closer.}
    \label{fig:similarity_example}
\end{figure}

Motivated by the above discussion, we argue that graph sparsification can greatly improve the performance of GNNs in tasks where the graph topology is significant. Moreover, we argue that sparsification that is based on local properties of a graph is more in-line with the prevailing methodology of GCNs, as opposed to approaches that rely on global topologies, such as spectral methods. Therefore, we introduce \methodname, a new algorithm for edges pruning based on locality sensitive hashing (LSH) \citep{shakhnarovich2008nearest}. As a result, pruned graphs are qualified with structure dynamics that alleviate the ability to distinguish between different graphs and mitigate the computation burden. More importantly, similar environments of the input graph result in similar environments in the sparsified graph while dissimilar environments result in dissimilar environments in the sparsified graph with high probability, as depicted in Figure \ref{fig:similarity_example}. Consequently, we preserve the ability to distinguish between environments of the graph that were distinguishable prior to the sparsification process.

\section{Related Work}\label{section:related_work}
\paragraph{Graph Convolutional Networks (GCNs)} were introduced by \cite{kipf2016semi} and has been widely adopted for solving tasks involving graphs, such as node classification \citep{bhagat2011node, zhang2018network}, link prediction \citep{kumar2020link}, graph classification \citep{kriege2020survey, cai2018comprehensive}, and more. 
The most basic form uses simple aggregation functions to obtain a node representation as a function of its neighbors, such as average and summation. 
Later, the model was extended to more complex architectures which introduce more sophisticated aggregation functions. 
For instance, Graph attention Networks (GATs) \citep{velivckovic2017graph} use dot-products based attention to calculate weights for each edge. Additionally, \cite{corso2020principal} argued that the aggregation layers are unable to extract enough information from the nodes’ neighbourhoods in a single layer, resulting in a limited expressive power and learning abilities. 
To address this, they proposed Principal Neighbourhood Aggregation for Graph Nets (PNA) which combines multiple aggregators to improve the performance of the graph neural networks.

\paragraph{Graph Sparsification} aims to approximate a graph on the same set of vertices and edges \citep{benczur2002randomized}. 
Much research has been done in this field and several algorithms were proposed, including spectral sparsifiers \citep{spielman2008spectral, calandriello2018improved, chu2020graph}, sampling via Metropolis algorithms \citep{4781123}, and others \citep{sadhanala2016graph}.
In the context of training graph neural networks, the motivation is twofold: 1. accelerate training by performing fewer message-passing operations, and 2. regularization. For the former, \cite{srinivasa2020fast} propose FastGAT, a spectral sparsifier based on effective-resistance \citep{spielman2008graph} for acceleration of GATs. Specifically, they aim to accelerate GATs by constructing a sampled graph with far fewer edges for each attention head within the network. \cite{chen2018fastgcn} proposed FastGCN which samples vertices and edges through importance sampling and thus reduces the graph size. Another line of works \citep{rong2019dropedge, chen2020label, hasanzadeh2020bayesian, zheng2020robust, luo2021learning, kim2020find, ye2021sparse} perform graph sparsification to achieve regularization, with the aim to prevent over-smoothing and overfitting \citep{li2018deeper}.

\paragraph{Locality-sensitive hashing (LSH)} is a widely used technique for finding pairs of similar items in a large set efficiently. It has been successfully applied for accelerating the Transformer model \citep{vaswani2017attention} by mitigating the burden of computing pair-wise attentions as demonstrated by the Reformer \citep{kitaev2020reformer}. This is done by approximation of attention computations based on locality-sensitive hashing, which replaces the \text{$\mathcal{O}(L^2)$} factor in attention layers with \text{$\mathcal{O}(L\cdot log L)$}.

\section{Preliminaries}\label{section:preliminaries}
\subsection{Graph Convolutional Networks (GCNs)}
A graph is a 2-tuple \text{$G=(V,E)$} where $V$ is a set of nodes and \text{$E\subseteq V\times V$} is a set of undirected edges connecting pairs of nodes. We consider the setting in which each node $v$ is associated with a feature vector \text{$x_v\in \Real^d$}. The adjacency matrix of $G$, denoted by \text{$A\in \{0,1\}^{|V|\times |V|}$}, associates each edge \text{$e_{i,j}=(v_i,v_j)\in E$} with an entry $A_{i,j}$ indicating that $e_{i,j}\in E$. The number of edges connected to a node $v_i$ (also known as the degree of $v_i$) is denoted by $d_i = \sum_{j}{A_{i,j}}$. We define $D$ as a diagonal matrix where $D_{i,i}=d_i$ and $0$ elsewhere. For what follows, we consider the Graph Convolutional Network (GCN) model from \cite{kipf2016semi}. This model is recursively defined as
\begin{equation}\label{eq:gcn}
    H^{(l+1)}=\sigma(\Tilde{D}^{-\frac{1}{2}} \Tilde{A} \Tilde{D}^{-\frac{1}{2}}H^{(l)}W^{(l)}),
\end{equation}
where \text{$\Tilde{A}=A+I$} is the adjacency matrix of the undirected graph $G$ with additional self-loops, obtained by adding the identity matrix $I$, $W^{(l)}$ is a weight matrix associated with the $l^{th}$ layer of the model, and $\sigma(\cdot)$ is some non-linear activation function. The notation \text{$H^{(l)}$} relates to the result of the $l^{th}$ layer of the model, which is also the input to the \text{$l+1$} layer. The input to the first layer is $H^{(0)}=[x_1,\hdots,x_{|V|}]^{\intercal}\in \Real^{|V|\times d}$. We intend to accelerate the training and inference of this model by reducing the input size.

\subsection{Locality Sensitive Hashing (LSH)}

The Locality Sensitive Hashing algorithm is designed to approximate distances between vectors in a high-dimensional space in a lower dimensional space.
It is widely applicable in fields like data mining and machine learning, and is particularly useful for tasks such as approximate nearest neighbors.
This algorithm relies on the existence of a family $\mathcal{H}$ of hash functions mapping $\Real^{d}$ to some universe $U$. The definition of LSH is as follows.

\begin{definition}
A family of hash functions $\mathcal{H}$ is called $(R,c,P_1,P_2)$-sensitive if for any $p,q\in \Real^{d}$
\begin{itemize}
    \item $d(p,q) \leq R \Rightarrow P_{h \in_R \mathcal{H}}[h(q)=h(p)] \geq P_1$,
    \item $d(p,q) \geq cR \Rightarrow P_{h \in_R \mathcal{H}}[h(q)=h(p)] \leq P_2$.
\end{itemize}
\end{definition}
An interesting hash functions family satisfies the constraint that $P_1 > P_2$. 
In our work, we utilize LSH by mapping feature vectors of graph edges into integers, namely ``buckets'', so that similar features are mapped into the same buckets. 
Then, the hash of a set of edges, consisting of edges in a neighborhood of a node, is the bucket with the minimal hash value, which is referred as \minhash. When applied to numerous sets, this process results with the same minimal buckets for similar edge sets. Consequently, by choosing to keep edges that correspond to the \minhash\ values, we aim to preserve the similarities of different graph regions.

\section{\methodname}\label{section:methodology}
This section introduces the methodology of Locality Sensitive Pruning (LSP). 
Additionally, we provide a complexity analysis of the algorithm and address its limitations. Finally, we discuss the advantages of pruning edges from graphs and the prominent advantages of LSP over other methods.

\subsection{Overview}

Our aim is to modify the topology of a graph so that similar regions result in a similar, yet consistently sparsified structure. For this purpose, we base our approach on locality sensitive hashing that assigns signatures to edges so that similar edges are assigned with the same signature. Then, we specify \minhash\ as a rule for choosing which edges to keep;  that is, the edges that make up the sparsified graph. In \minhash, the selected edges correspond to signatures that are mapped to minimal hash values using pseudo-random hash functions. These psuedo-random hash functions are used across all graph regions, and thus similar regions of the original graph result in similar regions in the resulting sparsified graph.

The procedure of sparsifying a graph using LSP is as follows. Given a graph \text{$G=(V,E)$}, LSP constructs a new graph \text{$G'=(V,E')$} where \text{$E' \subset E$}. 
We assume every edge \text{$(i,j)\in E$} is associated with an attribute \text{$e_{i,j}\in \Real^{d}$}, and the set of $k$ pseudo-random functions \text{$\mathH=\{h_z|\forall z\in [k]:h_z:\Real^{d}\to \Real\}$} maps $d$-dimensional vectors into scalars under the locality-sensitive-hashing criteria (see Section~\ref{section:hashing} for more details on the choice of these functions).
For each node \text{$u \in V$}, we compute the hash values for all the edges connecting $u$ with its neighbors \text{$\neighborhood_u=\{v\in V|(u,v)\in~E\}$}. 
Then, we base our choice of edges using the \minhash\ rule, that is, edges whose hashes are mapped to the minimal value among all the other edges.  The overview of our sparsification method is described in Algorithm \ref{algo:sparse}.
\begin{algorithm}[!ht]
    \caption{\methodname}
    \label{algo:sparse}
    \begin{algorithmic}
    \State \textbf{Input:} $G=(V,E)$ \Comment{The input graph}
    \State \hspace{2.8em} $\mathH=\{h_1,...,h_k\}$ \Comment{A set of $k$ locality-sensitive hashing (LSH) functions mapping edge attributes to scalars}
    \State \textbf{Output:} A sparsified graph $G'=(V,E')$
    \State $E'\gets \emptyset$
    \For{$u\in V$}
        \State $E_u \gets \emptyset$
        \For{$i\in [k]$}
            \State $v_{min} \gets \argmin_{v\in \neighborhood_u}{h_i(e_{u,v})}$ \Comment{Apply \minhash}
            \State $E_u \gets E_u \cup \{(u, v_{min})\}$
        \EndFor
        \State $E' \gets E' \cup E_u$
    \EndFor
    \State \Return $G'=(V,E')$
    \end{algorithmic}
\end{algorithm}

Note that LSP assumes the existence of edge attributes. However, many real-world problems incorporate only node attributes. In these cases, we construct edge attributes from the nodes connected to its ends, as we describe in Appendix \ref{app:edge_attrs}.

\subsection{The Choice of LSH Functions}\label{section:hashing}
The description of LSP does not define a specific family of hash functions. This is because different datasets contain attributes with different nature. Consequently, this requires choosing appropriate hashing methodologies for each dataset. In the following, we present two common choices for hash functions that we use in our experiments (Section \ref{section:experiments}). We refer the readers to \citep{pauleve2010locality} for a review of other LSH function choices.

\paragraph{Binary signatures}
A LSH function that uses binary signatures is a composition \text{$h=h_2 \circ h_1$} of a binary signature mapping \text{$h_1:\Real^{d}\to \{0,1\}^{d}$} and a hashing functions \text{$h_2:\{0,1\}^{d}\to \mathbb{N}$} that maps binary signatures to integers. Specifically, we define $h_1$ as a thresholding function which is associated with a random vector $w$ (with each \text{$w_i\sim N(0,1)$}) that determines the threshold for each entry of an input vector $x$.
Formally, the thresholding is performed by comparing the entries of $x$ and $w$:
\begin{equation}
    h_1(x)_i = 
    \begin{cases}
        1,& x_i > {w}_i\\
        0,& \text{otherwise}
    \end{cases}.
\end{equation}
Then, $h_2$ maps the result to integers using the well-known message-digest algorithm \citep{rivest1992rfc1321} and bin it into one of $m$ bins by finding the remainder of the division by $m$. In our experiments (which will be presented in Section \ref{section:experiments}), we refer to this method as \text{$LSP-T$}.

\paragraph{Random Projections}
A LSH hashing function that uses random projections \citep{shakhnarovich2008nearest} aims to project inputs so that similar vectors result in proximate projections, and their quantization result in the same values. Let $w$ be some random vector with each $w_i\sim N(0,1)$. This random projection divides input vectors into bins of length $l$ using the hash function defined by
\begin{equation}
    h_{w,b}(x)=\left\lfloor \frac{\langle x,w\rangle+b}{l}\right\rfloor.
\end{equation}
Here, $b$ is a random variable sampled from the uniform distribution $b\sim U[0,l]$. The inner-product $\langle x,w\rangle$ is the projected value of $x$ onto the direction $w$, which is then shifted by the offset value $b$ and quantized into $l$ length bins. In our experiments, we refer to this method as \text{$LSP-P$}.

\subsection{Analysis}\label{section:analysis}
\paragraph{Complexity}
The procedure of LSP is performed as a preprocessing step prior to the whole training phase due to its determinism.
The pruned graph does not change between training episodes and thus should be performed only once. 
Assuming the dimensionality of each edge attribute is $d$, the complexity of computing each hash value is $\mathcal{O}(d)$ given that the hash function families proposed in Section \ref{section:hashing} are used (note that this statement is valid for other common hash function families). Each of the $k$ hash functions computers $|E|$ hash values, one for each edge. In total, the running time complexity of LSP is \text{$\mathcal{O}(k|E|d)$}. The space complexity is \text{$\mathcal{O}(1)$}, because this only requires iterating over edges and keeping the minimal hash value, which requires a constant amount of memory.

\paragraph{Limitations}
We base LSP on the assumption that the topology of the input graph plays a significant role in the learning procedure. For this reason, in problems where the graph topology is less meaningful, we do not expect a significant advantage of pruning using LSP over its rival, such as random removal of edges.

\subsection{Benefits of using LSP}
In the following, we present the beneficial reasons for pruning edges. While many claims are valid in general, that is, they are applied to other pruning methods other than LSP, we begin with advantages that are particular for LSP.

\subsubsection{Advantages of pruning using local environmental characteristics}\label{section:lsp_advantages}
Several graph-related tasks, including node-classification among them, rely on the composition of the neighborhoods. This results from the core methodology of GCNs which aggregates information from a node's local environment. Consequently, we want our pruned graph to preserve local properties of the original graph. For demystifying the necessity to specifically consider local properties for pruning, we exemplify a node classification scenario in which consistent pruning that preserves similarities between neighborhoods is necessary for successfully predicting the node classes.

\begin{figure}[!h]
    \centering
    \includegraphics[width=0.75\linewidth]{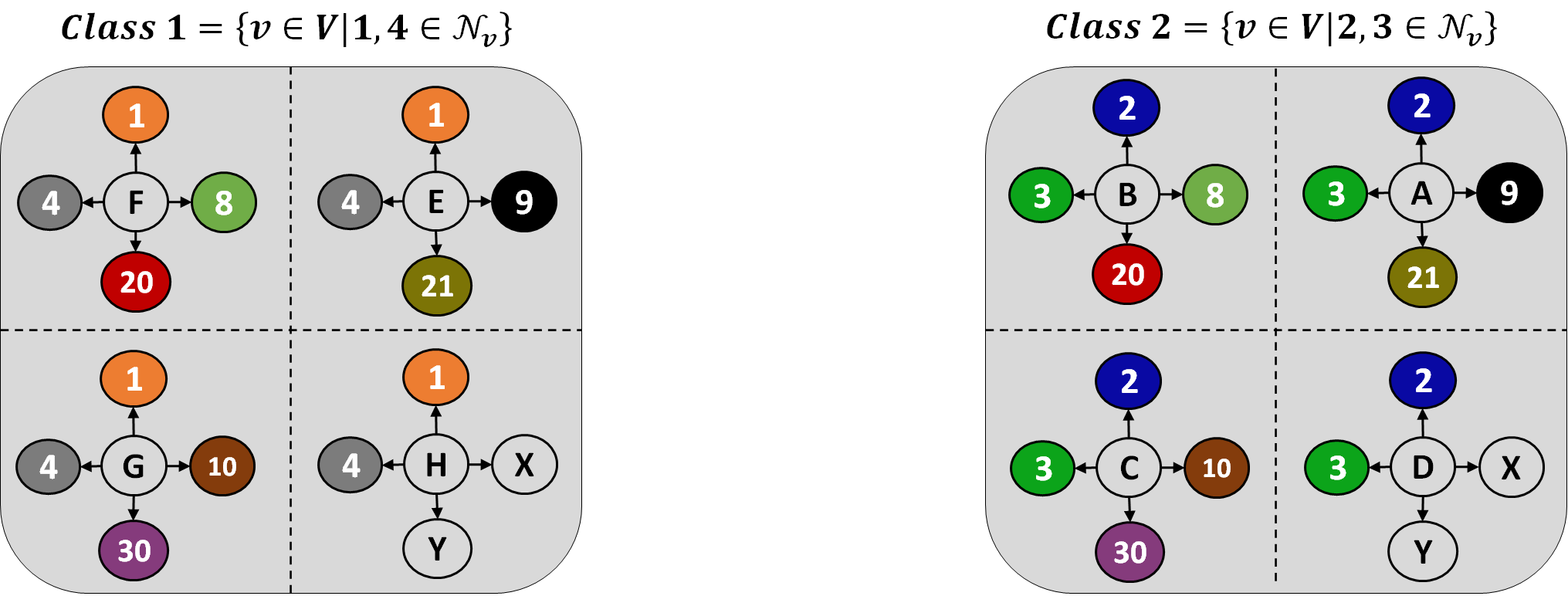}
    \caption{Example scenario in which we are tasked to classify nodes into 2 classes. Each of the nodes $A,B,C,D$ belong to class 2, while each of the nodes $E,F,G,H$ belong to class 1. Moreover, For each node among $\{E,F,H,G\}$ there exists a node among $\{A,B,C,D\}$ that the intersection of the sets of their neighbors has size of at least 2.}
    \label{fig:example:similarities}
\end{figure}

\begin{example}
This example is illustrated in Figure \ref{fig:example:similarities}.
Consider a node classification scenario consisting of 2 classes. The criterion for a node to belong to class 1 is being a neighbor of both nodes $1,4\in V$. Similarly, the criterion for a node to belong to class 2 is being a neighbor of both nodes $2,3\in V$. Notice that for each environment on the left, there's a corresponding environment on the right which has at least 2 common neighbors. For example, the top left environment contains nodes $8$ and $20$ as neighbors of $F$, while on the right we have nodes $8$ and $20$ as neighbors of $B$. 

Suppose that we want to reduce the degree of every node by $50\%$. When performing consistent pruning by choosing to keep the nodes corresponding to minimal identity (in our algorithm, these are the hash values) we encounter the following situation: on the left, we are left with environments consisting of nodes $E,F,G,H$, each of which has two neighbors $1$ and $4$. On the right, we are left with environments consisting of nodes $A,B,C,D$, each of which has two neighbors $2$ and $3$. These results will enable us to train a classifier that distinguish between the depicted components.

Meanwhile, suppose that we randomly prune $50\%$ of the edges of each node. Since there are $\binom{4}{2}=6$ possibilities, the probability that we are left with environments consisting of nodes $8$ and $20$ both on the left and the right is $\left(\frac{1}{6}\right)^2=\frac{1}{36}$. Consequently, a classifier trained on randomly pruned graphs using the described settings would have an error rate of $2.7\%$.
\end{example}

\subsubsection{Other Advantages of Pruning Edges of a Graph}\label{section:general_advantagres}
\paragraph{Accelerating GCN training and inference}
The outcome of pruning edges from a graph is an acceleration of each training iteration. To see this, recall the recursive definition of GCN (Equation \ref{eq:gcn}) from \cite{kipf2016semi}. In this definition, the weights matrix $W$ is of size \text{$C\times F$} where $C$ and $F$ denote the number of input channels and filters respectively. Additionally, the matrix $D$ is a diagonal matrix of size \text{$|V|\times |V|$} with at most $|V|$ non-zero elements, and the matrix $A$ is of size \text{$|V|\times|V|$} whose number of non-zero elements is at most $|E|$. Apparently, the complexity of this matrices multiplication does not depend on the number of edges, however from \cite{kipf2016semi} we know that in practice $\Tilde{A}H$ is efficiently implemented as a product of a sparse matrix with a dense matrix whose complexity is linear in the number of edges \text{$\mathcal{O}(|E|FC)$}. For GCNs that uses an attention mechanism, from \cite{velivckovic2017graph} we know that the time complexity of a single attention head is \text{$\mathcal{O}(|V|CF+|E|F)$}. With this, we claim that pruning edges from an input graph linearly accelerates GCNs and their variants.

\paragraph{Mitigating Over-Squashing}
LSP simplifies the graph structure by removing edges. In turn, it reduces the amount of information needed to be encoded within the node representation vectors. Particularly, pruning of edges helps mitigating the \textit{over-squashing} phenomenon in which information from the exponentially-growing receptive field is compressed into fixed-length node vectors \citep{alon2020bottleneck}. Since removing edges reduces the degree of each node, it also reduces its receptive field exponentially, thus preventing this destructive phenomenon.

We further discuss the \textit{Neighborhood Variance} issue in Appendix \ref{app:neighborhood_variance}.

\section{Experiments}\label{section:experiments}
We validated LSP on 3 graph-related tasks: (1) Node classification; (2) Graph regression, and (3) Graph classification. The objectives of this empirical study were twofold:
\begin{itemize}
    \item Assess the quality of the pruned graphs in terms of performance metrics. We train and test several models on various datasets and configurations, and measure the achieved performance metric identified with each dataset as a dependent of the pruning configurations.
    \item As we target acceleration of GNNs, we assess the amount of acceleration obtained by training and testing with pruned graphs. This is done by measuring the relative time required for training and testing on graphs that are pruned with various pruning configurations.
    
\end{itemize}

For all the experiments discussed in this section, for each dataset we choose a commonly used model that is identified with this dataset. For example, we use the GAT \citep{velivckovic2017graph} architecture proposed for the PPI \citep{zitnik2017predicting} dataset because it achieves state-of-the-art performance at the time of writing this paper. We refer the readers to Appendix \ref{app:models} for more information about the models that we use for experimenting with each dataset. Additional implementation details are given in Appendix \ref{app:experiments_details}.

\paragraph{Evaluation Protocol} To evaluate the quality of the pruned graphs, we train and test models with various pruning configurations and evaluate their performance in terms of the metric identified with each dataset. Additionally, we measure the running times in order to demonstrate the relative acceleration achieved via pruning. These results are presented as the relative time required for each iteration compared to the model trained on the unpruned graph. It is important to note the observation that given a pruning ratio, the choice of pruning methodology does not impact the running times, thus we present a single curve that describes the acceleration for all pruning methods.


\paragraph{Baselines} We compare the performance of 2 LSP variants (see Section \ref{section:hashing}) with \textit{Random}, that relates to the method which randomly removes edges from a graph with a certain probability $p$. We do not compare LSP with methods that sparsify the input graph during inference, such as FastGCN \citep{chen2018fastgcn} and DropEdge \citep{rong2019dropedge}, which are discussed in Section \ref{section:related_work}. This is because LSP does not modify the training and inference phases of GNNs, as opposed to the aforementioned methods. Additionally, in order to demonstrate the advantage of using the combination of LSP with complex models over using simpler GNNs which are computationally cheaper in terms of acceleration, we compare the performance of these models with various pruning configurations against a basic GCN that is naturally computationally cheaper (described in Appendix \ref{app:models}). 

\subsection{Node Classification Benchmarks}
\paragraph{Data}
We experimented with various publicly available datasets: GitHub \citep{rozemberczki2019multi}, Cora \citep{bojchevski2017deep}, CiteSeer \citep{bojchevski2017deep}, PubMed \citep{bojchevski2017deep} and PPI \citep{zitnik2017predicting}. Detailed descriptions and statistics of these datasets are presented in Appendix \ref{app:datasets}.

\paragraph{Results}
Results for node classification benchmarks are shown in Figure \ref{fig:node_classification_results}. Additional results are provided in Appendix \ref{app:additional_results}.
The superiority of LSP is demonstrated through the experiments conducted on Cora and PPI. For these benchmarks, the predictive performance drops significantly for aggressively pruned graphs via random, while it is noticeable that LSP better preserves the accuracy of the model trained on the original graphs. Moreover, the basic models used for these experiments demonstrate a significant decrease in performance while the acceleration achieved by their simplicity is negligible.
Notice that for PubMed and CiteSeer, the accuracy ranges are narrow. For instance, the model trained on PubMed using 100\% of the edges achieves an accuracy score of 0.882, while the same model that is trained with 30\% of the edges using random pruning achieves an accuracy score of 0.859. Considering this relatively negligible difference, we observe that random pruning performs well and consistent pruning via either LSP-P or LSP-T preserves this performance. Additionally, for this experiment the basic model demonstrates performance that is comparable to the original model while being substantially faster. The same holds for CiteSeer.

To that end, we conclude that in cases where random pruning performs well, the graph topology has a negligible impact on the performance of the model. In these cases LSP performs marginally better than random. The superiority of LSP is well-demonstrated in benchmarks where the topology of the graph is impactful. In such cases, the performance of the models reduces significantly via random pruning, while pruning via an LSP variant preserves the local properties of the graph. To conclude this experiments series, we observe that the achieved inference acceleration is linear in the number of preserved, a result that matches our expectation as per our complexity expectations.
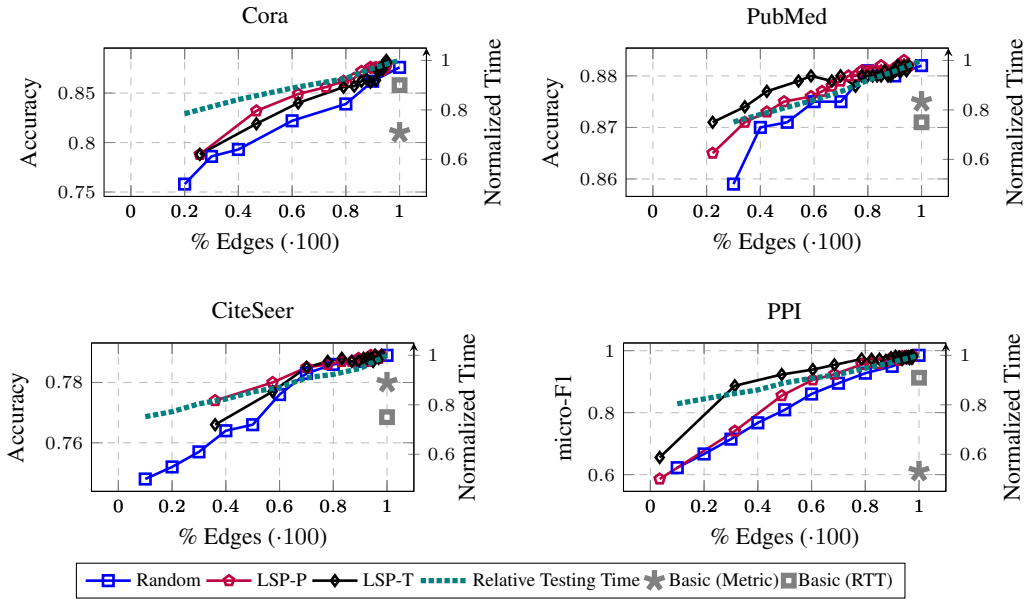
\begin{figure}[!ht]
    \centering
    \begin{subfigure}[b]{0.42\textwidth}
         \centering
         \begin{tikzpicture}
\small
    \begin{axis}[
        height=1.4in,
        width=\linewidth,
        title={Cora},
        xlabel={\% Edges ($\cdot 100$)},
        ylabel={Accuracy},
        xmin=0, xmax=1,
        xtick=\myxticks,
        xticklabel style={font=\scriptsize},
        yticklabel style={font=\scriptsize},
        ymajorgrids=true,
        xmajorgrids=true,
        grid style=dashed,
        enlargelimits=0.1,
        legend pos=south west,
        label style={font=\footnotesize},
    ]
     
    \addplot[
        color=blue,mark=square,style=solid,line width=1pt
        ]
        coordinates {
(0.2, 0.758)
(0.3, 0.786)
(0.4, 0.793)
(0.6, 0.822)
(0.8, 0.839)
(0.9, 0.862)
(1.0, 0.876)
        };
        
        \addplot[
        color=purple,mark=pentagon,style=solid,line width=1pt
        ]
        coordinates {
(0.2565365668813945, 0.788)
(0.4671276998863206, 0.832)
(0.6224895793861311, 0.849)
(0.7254641909814323, 0.856)
(0.7915877226222053, 0.862)
(0.8310913224706328, 0.865)
(0.8581849185297461, 0.872)
(0.8914361500568397, 0.876)
(0.9029935581659719, 0.875)
(0.9120879120879121, 0.876)
(0.9259189086775294, 0.876)
(0.9315081470253884, 0.873)
(0.9409814323607427, 0.874)
(0.9447707464948845, 0.876)
(0.9505494505494505, 0.878)
(0.9528230390299356, 0.880)
        };
        
        \addplot[
        color=black,mark=diamond,style=solid,line width=1pt
        ]
        coordinates {
(0.2565365668813945, 0.788)
(0.4671276998863206, 0.819)
(0.6224895793861311, 0.84)
(0.7915877226222053, 0.856)
(0.8310913224706328, 0.857)
(0.8581849185297461, 0.862)
(0.8775104206138689, 0.864)
(0.8914361500568397, 0.861)
(0.9120879120879121, 0.863)
(0.9198560060629026, 0.872)
(0.9259189086775294, 0.874)
(0.9365289882531261, 0.874)
(0.9409814323607427, 0.879)
(0.9447707464948845, 0.880)
(0.9478021978021978, 0.881)
(0.9505494505494505, 0.883)
(0.9528230390299356, 0.881)
        };
        
        \addplot[color=gray,mark=star,style=only marks,line width=2pt, mark options={scale=2}]
        coordinates {
(1.0, 0.81)
};
        
    \end{axis}
    
    \begin{axis}[
        height=1.4in,
        width=\linewidth,
        axis y line=right,
        xmin=0, xmax=1,
        xticklabel style={font=\scriptsize},
        yticklabel style={font=\scriptsize},
        ymin=0.5, ymax=1,
        ylabel={\timelabel},
        ylabel style={yshift=0pt},
        xtick=\myxticks,
        enlargelimits=0.1,
        label style={font=\footnotesize},
    ]
        \addplot[color=teal, style=densely dotted, line width=2pt]
          coordinates{
(0.2, 0.784687085249525)
(0.3, 0.8134478877744048)
(0.4, 0.8417883727749748)
(0.6, 0.8883096448777311)
(0.8, 0.9271812509542506)
(0.9, 0.9689970163695954)
(1.0, 1.0)
        };
        
        \addplot[
        color=gray,mark=square,style=only marks,line width=2pt
        ]
        coordinates {
(1.0, 0.9)
        };

    \end{axis}
    \end{tikzpicture}
     \end{subfigure}\qquad\qquad\hspace{-12pt}
    \begin{subfigure}[b]{0.42\textwidth}
         \centering
         \begin{tikzpicture}
\small
    \begin{axis}[
        height=1.4in,
        width=\linewidth,
        title={PubMed},
        xlabel={\% Edges ($\cdot 100$)},
        ylabel={Accuracy},
        xmin=0, xmax=1,
        xtick=\myxticks,
        xticklabel style={font=\scriptsize},
        yticklabel style={font=\scriptsize},
        ymajorgrids=true,
        xmajorgrids=true,
        grid style=dashed,
        enlargelimits=0.1,
        legend pos=south west,
        label style={font=\footnotesize},
    ]
     
    \addplot[
        color=blue,mark=square,style=solid,line width=1pt
        ]
        coordinates {
(0.3, 0.859)
(0.4, 0.87)
(0.5, 0.871)
(0.6, 0.875)
(0.7, 0.875)
(0.8, 0.881)
(0.9, 0.88)
(1.0, 0.882)
        };
        
        \addplot[
        color=purple,mark=pentagon,style=solid,line width=1pt
        ]
        coordinates {
(0.22241900550491833, 0.865)
(0.34225250428661674, 0.871)
(0.4242171284180128, 0.873)
(0.4883133291219204, 0.875)
(0.588642721776013, 0.876)
(0.6296475949824023, 0.877)
(0.6658807869325873, 0.878)
(0.6985493186535511, 0.879)
(0.727799837559787, 0.880)
(0.7540610053244292, 0.880)
(0.7777389224799206, 0.881)
(0.7988561501669524, 0.881)
(0.8177962277772765, 0.881)
(0.8345704358812381, 0.881)
(0.8494607887374785, 0.882)
(0.8628282645970581, 0.881)
(0.8746954246006678, 0.881)
(0.8852427578738381, 0.881)
(0.894684595253136, 0.881)
(0.9032916704268568, 0.881)
(0.9110301416839636, 0.882)
(0.924115603284902, 0.882)
(0.9346854976987636, 0.883)
(0.9438566013897662, 0.882)
(0.9516965977799837, 0.882)
        };
        
        \addplot[
        color=black,mark=diamond,style=solid,line width=1pt
        ]
        coordinates {
(0.22241900550491833, 0.871)
(0.34225250428661674, 0.874)
(0.4242171284180128, 0.877)
(0.5420990885299161, 0.879)
(0.588642721776013, 0.880)
(0.6658807869325873, 0.879)
(0.6985493186535511, 0.880)
(0.7540610053244292, 0.878)
(0.7777389224799206, 0.880)
(0.7988561501669524, 0.880)
(0.8177962277772765, 0.880)
(0.8345704358812381, 0.880)
(0.8494607887374785, 0.880)
(0.8628282645970581, 0.881)
(0.8746954246006678, 0.880)
(0.8852427578738381, 0.880)
(0.894684595253136, 0.881)
(0.9032916704268568, 0.881)
(0.9110301416839636, 0.882)
(0.9179564118761845, 0.881)
(0.924115603284902, 0.881)
(0.9346854976987636, 0.882)
(0.9438566013897662, 0.881)
(0.9516965977799837, 0.882)
        };
        
        \addplot[color=gray,mark=star,style=only marks,line width=2pt, mark options={scale=2}]
        coordinates {
(1.0, 0.875)
};
        
    \end{axis}
    
    \begin{axis}[
        height=1.4in,
        width=\linewidth,
        axis y line=right,
        xmin=0, xmax=1,
        ymin=0.5, ymax=1,
        xticklabel style={font=\scriptsize},
        yticklabel style={font=\scriptsize},
        ylabel={\timelabel},
        xtick=\myxticks,
        enlargelimits=0.1,
        label style={font=\footnotesize},
    ]
        \addplot[color=teal, style=densely dotted, line width=2pt]
          coordinates{
(0.3, 0.7511379403472075)
(0.4, 0.7814919471698336)
(0.5, 0.8108158239275119)
(0.6, 0.8398974989334676)
(0.7, 0.8745488132951173)
(0.8, 0.919639126802277)
(0.9, 0.9564726358219889)
(1.0, 1.0)
        };
        
        \addplot[
        color=gray,mark=square,style=only marks,line width=2pt
        ]
        coordinates {
(1.0, 0.75)
        };
        
        

    \end{axis}
    \end{tikzpicture}
     \end{subfigure}
     \begin{subfigure}[b]{0.42\textwidth}
         \centering
         \begin{tikzpicture}
\small
    \begin{axis}[
        height=1.4in,
        width=\linewidth,
        title={CiteSeer},
        xlabel={\% Edges ($\cdot 100$)},
        ylabel={Accuracy},
        xmin=0, xmax=1,
        xtick=\myxticks,
        xticklabel style={font=\scriptsize},
        yticklabel style={font=\scriptsize},
        ymajorgrids=true,
        xmajorgrids=true,
        grid style=dashed,
        enlargelimits=0.1,
        legend pos=south west,
        label style={font=\footnotesize},
    ]
     
    \addplot[
        color=blue,mark=square,style=solid,line width=1pt
        ]
        coordinates {
(0.1, 0.748)
(0.2, 0.752)
(0.3, 0.757)
(0.4, 0.764)
(0.5, 0.766)
(0.6, 0.776)
(0.7, 0.783)
(0.8, 0.786)
(0.9, 0.787)
(1.0, 0.789)
        };
        
        \addplot[
        color=purple,mark=pentagon,style=solid,line width=1pt
        ]
        coordinates {
(0.3601713532513181, 0.774)
(0.5741432337434095, 0.780)
(0.7007908611599297, 0.785)
(0.7793277680140598, 0.786)
(0.8318321616871704, 0.787)
(0.8690685413005272, 0.787)
(0.8942223198594025, 0.788)
(0.9126757469244289, 0.787)
(0.9275043936731108, 0.787)
(0.9396968365553603, 0.789)
(0.9485940246045694, 0.788)
(0.9554042179261862, 0.789)
(0.9690246045694201, 0.788)
(0.9798989455184535, 0.789)
        };
        
        \addplot[
        color=black,mark=diamond,style=solid,line width=1pt
        ]
        coordinates {
(0.3601713532513181, 0.766)
(0.5741432337434095, 0.777)
(0.7007908611599297, 0.785)
(0.7793277680140598, 0.787)
(0.8318321616871704, 0.788)
(0.8690685413005272, 0.787)
(0.8942223198594025, 0.787)
(0.9126757469244289, 0.788)
(0.9275043936731108, 0.788)
(0.9396968365553603, 0.788)
(0.9485940246045694, 0.787)
(0.9554042179261862, 0.789)
(0.9690246045694201, 0.788)
(0.9798989455184535, 0.789)
        };
        
        \addplot[color=gray,mark=star,style=only marks,line width=2pt, mark options={scale=2}]
        coordinates {
(1.0, 0.78)
};
        
    \end{axis}
    
    \begin{axis}[
        height=1.4in,
        width=\linewidth,
        axis y line=right,
        xmin=0, xmax=1,
        ymin=0.5, ymax=1,
        xticklabel style={font=\scriptsize},
        yticklabel style={font=\scriptsize},
        ylabel={\timelabel},
        ylabel style={yshift=5pt},
        xtick=\myxticks,
        enlargelimits=0.1,
        label style={font=\footnotesize},
    ]
        \addplot[color=teal, style=densely dotted, line width=2pt]
          coordinates{
(0.1, 0.7522020594874795)
(0.2, 0.7721805566177885)
(0.3, 0.8040854565093087)
(0.4, 0.8232223847521216)
(0.5, 0.852340634596584)
(0.6, 0.8702750458955674)
(0.7, 0.909085750906387)
(0.8, 0.9228992443490338)
(0.9, 0.9470032720028777)
(1.0, 1.0)
        };
        
        \addplot[
        color=gray,mark=square,style=only marks,line width=2pt
        ]
        coordinates {
(1.0, 0.75)
        };
        
        

    \end{axis}
    \end{tikzpicture}
     \end{subfigure}\qquad\qquad
     \begin{subfigure}[b]{0.42\textwidth}
         \centering
         \begin{tikzpicture}
\small
    \begin{axis}[
        height=1.4in,
        width=\linewidth,
        title={PPI},
        xlabel={\% Edges ($\cdot 100$)},
        ylabel={micro-F1},
        xmin=0, xmax=1,
        xtick=\myxticks,
        xticklabel style={font=\scriptsize},
        yticklabel style={font=\scriptsize},
        ymajorgrids=true,
        xmajorgrids=true,
        grid style=dashed,
        enlargelimits=0.1,
        legend pos=south west,
        label style={font=\footnotesize},
    ]
     
    \addplot[
        color=blue,mark=square,style=solid,line width=1pt
        ]
        coordinates {
(0.0999922150968734, 0.6224774141738607)
(0.19999355312709827, 0.666730808670311)
(0.29999367476620964, 0.7145453991530502)
(0.39999379640532096, 0.7672059618109716)
(0.4999963508266594, 0.809194832396935)
(0.5999940396835437, 0.8601492495484041)
(0.6999929449315415, 0.8950939709500391)
(0.7999942829617664, 0.927439837369523)
(0.8999944046008778, 0.9495145972652588)
(1.0, 0.9850249166727275)
        };
        
        \addplot[
        color=purple,mark=pentagon,style=solid,line width=1pt
        ]
        coordinates {
(0.034633087784513884, 0.5858800046367371)
(0.3143635111859327, 0.7405198592940659)
(0.48898253748917414, 0.8549889296591583)
(0.6047933108219885, 0.9058696398377808)
(0.6849504442260347, 0.9239085112225285)
(0.7863165730856436, 0.9552369569841287)
(0.8233246645193308, 0.9616003991925423)
(0.8526524624621702, 0.9579118246288156)
(0.8763915514338818, 0.9656280941139718)
(0.8961481758998862, 0.9763553737445136)
(0.9125597248036744, 0.9741784979681933)
(0.9259856417192959, 0.9686411850630167)
(0.9368911962476767, 0.9813135752856964)
(0.9457878808520577, 0.9741899200088378)
(0.95319083716902, 0.9820575512507053)
(0.9594880939637808, 0.9794111291465444)
(0.9649521228457714, 0.9736480045475239)
(0.9695804910327647, 0.9819340966672431)
(0.9735185572628281, 0.9838775669006925)
(0.9768484279361249, 0.984267690069481)
        };
        
        \addplot[
        color=black,mark=diamond,style=solid,line width=1pt
        ]
        coordinates {
(0.034633087784513884, 0.6553892003122117)
(0.3143635111859327, 0.8868053490277104)
(0.48898253748917414, 0.923163176343815)
(0.6047933108219885, 0.9390102692608817)
(0.6849504442260347, 0.9542127659574469)
(0.7863165730856436, 0.9730358241565062)
(0.8233246645193308, 0.9721536320091017)
(0.8526524624621702, 0.9724858565712065)
(0.8763915514338818, 0.9682609759604832)
(0.8961481758998862, 0.9767862375372225)
(0.9125597248036744, 0.9817701864133314)
(0.9259856417192959, 0.9798376669838345)
(0.9368911962476767, 0.9717662706689641)
(0.9457878808520577, 0.9796929619989068)
(0.95319083716902, 0.97800039186331)
(0.9594880939637808, 0.9809830354058938)
(0.9649521228457714, 0.974986355050392)
(0.9695804910327647, 0.9820908142563023)
(0.9735185572628281, 0.9756411740138978)
(0.9768484279361249, 0.9821515898552214)
        };
        
        \addplot[color=gray,mark=star,style=only marks,line width=2pt, mark options={scale=2}]
        coordinates {
(1.0, 0.6103558304054036)
};
    \end{axis}
    
    \begin{axis}[
        height=1.4in,
        width=\linewidth,
        axis y line=right,
        xmin=0, xmax=1,
        ymin=0.5, ymax=1,
        xticklabel style={font=\scriptsize},
        yticklabel style={font=\scriptsize},
        ylabel={\timelabel},
        xtick=\myxticks,
        enlargelimits=0.1,
        label style={font=\footnotesize},
    ]
        \addplot[color=teal, style=densely dotted, line width=2pt]
          coordinates{
(0.0999922150968734, 0.8040211301667167)
(0.19999355312709827, 0.8238529309126918)
(0.29999367476620964, 0.8431458070846159)
(0.39999379640532096, 0.8603021434747149)
(0.4999963508266594, 0.8886452250699007)
(0.5999940396835437, 0.9073824668642728)
(0.6999929449315415, 0.92285965854122)
(0.7999942829617664, 0.947859808763143)
(0.8999944046008778, 0.971790678961688)
(1.0, 1.0)
        };
        
        \addplot[
        color=gray,mark=square,style=only marks,line width=2pt
        ]
        coordinates {
(1.0, 0.909459597246477)
        };
        

    \end{axis}
    \end{tikzpicture}
     \end{subfigure}
    
         \vspace{-1em}
     \begin{subfigure}[b]{\textwidth}
         \centering
         \begin{tikzpicture}
\small
    \begin{axis}[
        axis lines=none,
        height=0.8in,
        width=0.8in,
        title={},
        xmin=0, xmax=1,
        ymin=0, ymax=1,
        ytick={},
        xtick={},
        legend columns=6,
        legend style={font=\footnotesize, anchor=south east,at={(1,0)},nodes={scale=0.8, transform shape}},
    ]
     \addplot+[forget plot]
       coordinates{
   (0, 2.388854806)
       };
    \addlegendimage{color=blue,mark=square,style=solid,line width=1pt}
    \addlegendentry{Random}
    \addlegendimage{color=purple,mark=pentagon,style=solid,line width=1pt}
    \addlegendentry{LSP-P}
    \addlegendimage{color=black,mark=diamond,style=solid,line width=1pt}
    \addlegendentry{LSP-T}
    \addlegendimage{color=teal, style=densely dotted, line width=2pt}
    \addlegendentry{Relative Testing Time}
    \addlegendimage{color=gray,mark=star,style=only marks,line width=2pt, mark options={scale=2}}
    \addlegendentry{Basic (Metric)}
    \addlegendimage{color=gray,mark=square,style=only marks,line width=2pt}
    \addlegendentry{Basic (RTT)}
    \end{axis}
    \end{tikzpicture}
     \end{subfigure}
    \caption{Test results for node classification problems. For each dataset, we report the results in terms of the metric identified with that dataset (in solid lines), and the running time per iteration (in densely dotted lines) under the specified pruning configurations.}
    \label{fig:node_classification_results}
\end{figure}

\subsection{Graph Regression Benchmarks}
\paragraph{Data}
We experimented with datasets from the chemical world with the goal to predict chemical properties of molecules: QM9 \citep{wu2017moleculenet} and ZINC \citep{sterling2015zinc}. In both datasets, each graph is a molecule, consisting of nodes and edges that represent atoms and different types of bonds between the atoms respectively. The goal is to regress each graph to its properties, quantum properties in QM9 and constrained solubility (a synthetic computed property) in ZINC. Refer to Appendix \ref{app:datasets} for more details of these datasets.

\paragraph{Results}
Results for the graph regression tasks are shown in Figure \ref{fig:graph_regression_results}. We include additional results for other QM9 targets in Appendix \ref{app:additional_results}. Continuing the conclusions from the previous section, we observe that pruning via LSP perform better in terms of lower Mean Absolute Error (MAE) for both graph regression benchmarks.
\begin{figure}[!ht]
    \centering
    \begin{subfigure}[b]{0.42\textwidth}
         \centering
         \begin{tikzpicture}
\small
    \begin{axis}[
        height=1.4in,
        width=\linewidth,
        title={constrained solubility},
        xlabel={\% Edges ($\cdot 100$)},
        ylabel={MAE},
        xtick=\myxticks,
        xticklabel style={font=\scriptsize},
        yticklabel style={font=\scriptsize},
        ymajorgrids=true,
        xmajorgrids=true,
        grid style=dashed,
        enlargelimits=0.1,
        legend pos=south west,
        label style={font=\footnotesize},
    ]
     
    \addplot[
        color=blue,mark=square,style=solid,line width=1pt
        ]
        coordinates {
(0,  0.6716939210891724 )
(0.1,  0.6823663711547852 )
(0.2,  1.04264235496521 )
(0.3,  0.9010999798774719 )
(0.4,  0.8267825841903687 )
(0.5,  0.7396655082702637 )
(0.6,  0.6791566014289856 )
(0.7,  0.610236406326294 )
(0.8,  0.5403611660003662 )
(0.9,  0.44265732169151306 )
(1.0,  0.19980350136756897 )
        };
        
        \addplot[
        color=purple,mark=pentagon,style=solid,line width=1pt
        ]
        coordinates {
(0.46466810280849974, 0.6306300082702637)
(0.8468382816041463, 0.4003611216915130)
(0.9937740443438877, 0.19980350136756897)
        };
        
        \addplot[
        color=black,mark=diamond,style=solid,line width=1pt
        ]
        coordinates {
(0.46466810280849974, 0.5306300082702637)
(0.8468382816041463, 0.3203611216915130)
(0.9937740443438877, 0.19980350136756897)
        };
    \end{axis}
    
    \begin{axis}[
        height=1.4in,
        width=\linewidth,
        axis y line=right,
        xticklabel style={font=\scriptsize},
        yticklabel style={font=\scriptsize},
        ylabel={\timelabel},
        xtick=\myxticks,
        label style={font=\footnotesize},
        enlargelimits=0.1,
    ]
    \addplot[color=teal, style=densely dotted, line width=2pt]
      coordinates{
(0.0, 0.8227427844050025)
(0.09210563264454687, 0.9003196626120787)
(0.19203783712226058, 0.9177348707358471)
(0.29201015729363483, 0.9221113684419725)
(0.39194236177134856, 0.9319823882920768)
(0.5, 0.9465678705913897)
(0.5921056326445469, 0.9595561733231453)
(0.6920378371222605, 0.9643849460060686)
(0.7920101572936348, 0.9724301381135833)
(0.8919423617713486, 0.9835380631302613)
(1.0, 1.0)
        };

        

    \end{axis}
    \end{tikzpicture}
         \vspace{-1.2em}
         \caption{ZINC}
         \label{fig:zinc_results}
     \end{subfigure}\qquad\qquad
    \begin{subfigure}[b]{0.42\textwidth}
         \centering
         \begin{tikzpicture}
\small
    \begin{axis}[
        height=1.4in,
        width=\linewidth,
        title={$\mu$},
        xlabel={\% Edges ($\cdot 100$)},
        ylabel={MAE ($D$)},
        xtick=\myxticks,
        xticklabel style={font=\scriptsize},
        yticklabel style={font=\scriptsize},
        ymajorgrids=true,
        xmajorgrids=true,
        grid style=dashed,
        enlargelimits=0.1,
        legend pos=south west,
        label style={font=\footnotesize},
    ]
     
    \addplot[
        color=blue,mark=square,style=solid,line width=1pt
        ]
        coordinates {
(0.0, 0.033735979021088505)
(0.1, 0.031349508989965726)
(0.2, 0.03237473956413305)
(0.3, 0.030768171834866225)
(0.4, 0.03299352678628616)
(0.5, 0.031641624874995056)
(0.6, 0.03282199324140636)
(0.7, 0.03131845086611739)
(0.8, 0.03351728260013704)
(0.9, 0.033531075667383364)
(1.0, 0.0255)
        };
        
        \addplot[
        color=purple,mark=pentagon,style=solid,line width=1pt
        ]
        coordinates {
(0.48318212868875066, 0.029693041048877256)
(0.7035288004586339, 0.02807081115592106)
(0.8768405079558135, 0.0270357962469907)
(0.9994301909075598, 0.025589568202613155)
        };
        
        \addplot[
        color=black,mark=diamond,style=solid,line width=1pt
        ]
        coordinates {
(0.48318212868875066, 0.0285888435474105)
(0.7035288004586339, 0.027216341177427744)
(0.8768405079558135, 0.026321315510747508)
(0.9994301909075598, 0.025589568202613155)
        };
    \end{axis}
    
    \begin{axis}[
        height=1.4in,
        width=\linewidth,
        axis y line=right,
        xticklabel style={font=\scriptsize},
        yticklabel style={font=\scriptsize},
        ylabel={\timelabel},
        xtick=\myxticks,
        label style={font=\footnotesize},
        enlargelimits=0.1,
    ]
        \addplot[color=teal, style=densely dotted, line width=2pt]
          coordinates{

(0.0, 0.5229324280744425)
(0.1, 0.9033374756138199)
(0.2, 0.908055830409958)
(0.3, 0.9207097496247466)
(0.4, 0.9326186614300677)
(0.5, 0.9404050326687592)
(0.6, 0.93553133752168)
(0.7, 0.9629465579994054)
(0.8, 0.9759265067151759)
(0.9, 0.9803781473953123)
(1.0, 1.0)
        };

        

    \end{axis}
    \end{tikzpicture}
         \vspace{-1.2em}
         \caption{QM9}
         \label{fig:qm9_results_mu}
     \end{subfigure}
     
     \begin{subfigure}[b]{\textwidth}
         \centering
         \begin{tikzpicture}
\small
    \begin{axis}[
        axis lines=none,
        height=0.8in,
        width=0.8in,
        title={},
        xmin=0, xmax=1,
        ymin=0, ymax=1,
        ytick={},
        xtick={},
        legend columns=6,
        legend style={font=\footnotesize, anchor=south east,at={(1,0)},nodes={scale=0.8, transform shape}},
    ]
     \addplot+[forget plot]
       coordinates{
   (0, 2.388854806)
       };
    \addlegendimage{color=blue,mark=square,style=solid,line width=1pt}
    \addlegendentry{Random}
    \addlegendimage{color=purple,mark=pentagon,style=solid,line width=1pt}
    \addlegendentry{LSP-P}
    \addlegendimage{color=black,mark=diamond,style=solid,line width=1pt}
    \addlegendentry{LSP-T}
    \addlegendimage{color=teal, style=densely dotted, line width=2pt}
    \addlegendentry{Relative Testing Time}
    \end{axis}
    \end{tikzpicture}
     \end{subfigure}
    \caption{Mean Absolute Error (MAE) and normalized running times for graph regression problems. (\ref{fig:zinc_results}) results for regressing the constrained solubility for molecules in the ZINC database. (\ref{fig:qm9_results_mu}) results for regressing the Dipole moment ($\mu$) for molecules in the QM9 database. Errors are presented as solid lines (Lower is better). Normalized running times are presented as dashed lines.}
    \label{fig:graph_regression_results}
\end{figure}
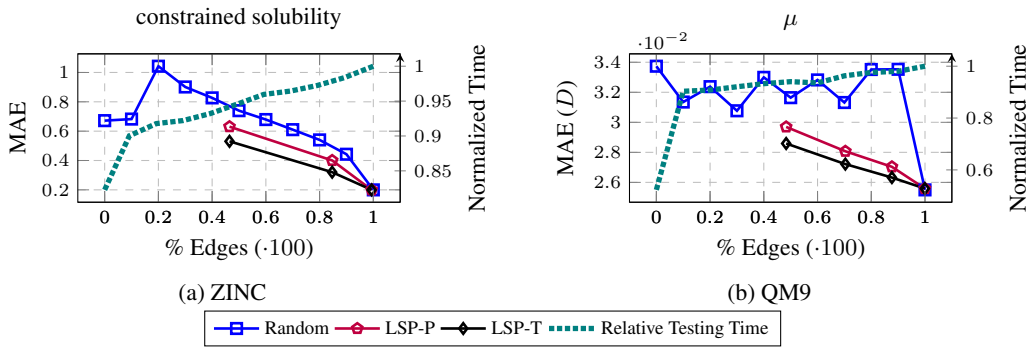

\subsection{Graph Classification Benchmarks}\label{section:graph_classification_experiments}

\begin{wrapfigure}{R}{0.5\linewidth}
\newcommand{\synHeight}{1.9in}

\begin{tikzpicture}
\begin{axis}[
    height=\synHeight,
    width=\linewidth,
    title={},
    xlabel={\% Edges ($\cdot 100$)},
    ylabel={Accuracy},
    ymin=0, ymax=1,
    xtick=\myxticks,
    ytick={0,0.2,0.4,0.6,0.8,1},
    xticklabel style={font=\scriptsize},
    yticklabel style={font=\scriptsize},
    minor y tick num=1,
    ylabel style={yshift=30pt},
    label style={font=\footnotesize},
    yminorgrids=true,
    ymajorgrids=true,
    xmajorgrids=true,
    grid style=dashed,
    enlargelimits=0.05,
    y label style={at={(axis description cs:0.1,.5)},anchor=south},
    legend style={anchor=south east,at={(1,0)},nodes={scale=0.1, transform shape}},
]
    \addplot[
    color=blue,
    mark=square,
    style=solid,
    line width=1pt
    ]
    coordinates {
(0.0, 0.2056)
(0.1, 0.245)
(0.2, 0.2668)
(0.3, 0.3198)
(0.4, 0.4116)
(0.5, 0.5598)
(0.6, 0.689)
(0.7, 0.7642)
(0.8, 0.7992)
(0.9, 0.8248)
(1.0, 0.912)
    }; \label{plot_one}
\addplot[
    color=purple,
    mark=pentagon,
    style=solid,
    line width=1pt
    ]
    coordinates {
(0.11257253867756076, 0.4532)
(0.3371431317877314, 0.626)
(0.5541228014592837, 0.7616)
(0.7431701793426628, 0.8524)
(0.8788702920058693, 0.8826)
(0.9544958255145907, 0.8904)
(0.986227322168625, 0.8914)
(0.9966995834803152, 0.8914)
(0.9993795612854259, 0.898)
(0.9999221758841597, 0.9076)
    }; \label{plot_two}
\addplot[
    color=black,
    mark=diamond,
    style=solid,
    line width=1pt
    ]
    coordinates {
(0.11257253867756076, 0.436)
(0.3371431317877314, 0.5984)
(0.5541228014592837, 0.7666)
(0.7431701793426628, 0.8588)
(0.8788702920058693, 0.8892)
(0.9544958255145907, 0.8943)
(0.986227322168625, 0.8946)
(0.9966995834803152, 0.892)
(0.9993795612854259, 0.8972)
(0.9999221758841597, 0.9066)
    }; \label{plot_three}
    
        \addplot[color=gray,mark=star,style=only marks,line width=2pt, mark options={scale=2}]
        coordinates {
(1.0, 0.34)
}; \label{plot_four}
    
\end{axis}

\begin{axis}[
    height=\synHeight,
    width=\linewidth,
    axis y line=right,
    ymin=0, ymax=1,
    ylabel={\timelabel},
    xticklabel style={font=\scriptsize},
    yticklabel style={font=\scriptsize},
    xtick=\myxticks,
    ytick={0,0.2,0.4,0.6,0.8,1},
    enlargelimits=0.05,
    legend columns=3,
    legend style={font=\footnotesize, anchor=south east,at={(1,0)},nodes={scale=0.66, transform shape}},
    label style={font=\footnotesize},
]
    \addlegendimage{/pgfplots/refstyle=plot_one}
    \addlegendentry{Random}
    \addlegendimage{/pgfplots/refstyle=plot_two}
    \addlegendentry{LSP-P}
    \addlegendimage{/pgfplots/refstyle=plot_three}
    \addlegendentry{LSP-T}
    
        \addplot[color=teal, style=densely dotted, line width=2pt]
          coordinates{
(0.0, 0.46457060706680864)
(0.1, 0.532176335031618)
(0.2, 0.5980282976238097)
(0.3, 0.6381016823062546)
(0.4, 0.68939937993684)
(0.5, 0.7346060584942004)
(0.6, 0.7951648240980242)
(0.7, 0.8389457022574128)
(0.8, 0.8797466653767281)
(0.9, 0.8899543633187307)
(1.0, 1.0)
        };
        \addlegendentry{RTT}
    
    \addlegendimage{/pgfplots/refstyle=plot_four}
    \addlegendentry{Basic (Metric)}
    
    \addplot[
    color=gray,mark=square,style=only marks,line width=2pt
    ]
    coordinates {
(1.0, 0.75)
    }; \addlegendentry{Basic (RTT)}
        

        \addplot[
        color=gray,style=solid,line width=2pt
        ]
        coordinates {
(0.55, 0.75)
(1.0, 0.75)
        };

    \end{axis}
\end{tikzpicture}
         \vspace{-1.2em}
\caption{Accuracy and running time comparison for the synthetic graph classification dataset.}\label{fig:graph_classification_comparison}
\end{wrapfigure}
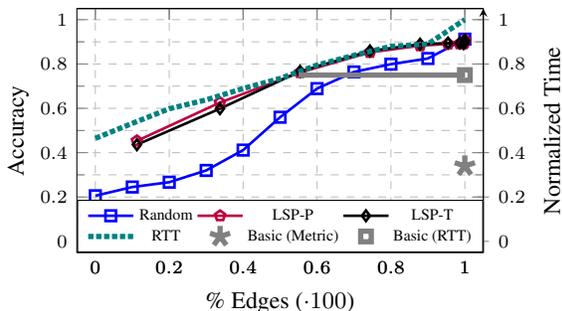

\paragraph{Data}
In order to validate the effectiveness of LSP in extreme scenarios, we developed a generator for synthetic graph classification datasets. We provide the full list of parameters and their default values of this generator in Appendix \ref{app:syn_generator}. We refer to the default generator configurations as those which we use in the following experiments.

\paragraph{Results}

Results for graph classification on the synthetic benchmark are shown in Figure \ref{fig:graph_classification_comparison}. This synthetic data generator enables us to simulate highly complex scenarios where the topology is impactful. This is well-demonstrated by the aggressive accuracy reduction when pruning with random. This yields the observation that LSP outperforms the random pruning method by a large margin, especially for mid-range pruning ratios. Moreover, as we use a highly complex model in this section (as described in Appendix \ref{app:models}) that performs complex calculations for neighborhood aggregation, a significant acceleration is observed when compared to the model trained on the original graphs. Additionally, this acceleration is depicted as a linear curve, which matches our expectations. The basic model used for this experiment (described in Appendix \ref{app:basic_models}) demonstrates a significant acceleration due to the lack of need to compute 16 attention heads per layer. This acceleration is also followed by a reduction of $\sim54\%$ in accuracy: the original model trained on the original graphs achieves an accuracy score of $90.76\%$ while the basic model achieves an accuracy score of $34\%$. For the sake of comparison, the original model trained on pruned graphs with the configuration that achieves comparable acceleration results in accuracy score of $\sim76\%$ for both LSP variants. In this configuration, nearly $45\%$ of the graph edges are pruned. From this result, we conclude that we were able to train a highly complex model that does not compensate for the performance while being accelerated via pruning its input.



\section{Conclusions}\label{section:conclusions}
We have presented Locality Sensitive Pruning (LSP), a systematic method for pruning graphs based on locality sensitive hashing. By relying on the local features of neighborhoods, LSP preserves local properties of the graph. Consequently, it does not impair the ability to distinguishing between samples that were distinguishable prior to the pruning process with high probability. We evaluated LSP through an extensive experimental study on 3 graph-related tasks. For these experiments, we trained several ad-hoc graph neural networks on diverse real-world datasets that went through the pruning process of 2 LSP variants. In all experiments, LSP demonstrates superior performance when compared to the baselines. At the same time, the reduction in the number of edges translates to a significant acceleration of the used models, which is linear in the number of edges. A possibly rewarding avenue of future research is the design of dedicated locality sensitive hash functions that capture informative features of the graph.
\newpage

\bibliography{iclr2022_conference}

\begin{thebibliography}{48}
\providecommand{\natexlab}[1]{#1}
\providecommand{\url}[1]{\texttt{#1}}
\expandafter\ifx\csname urlstyle\endcsname\relax
  \providecommand{\doi}[1]{doi: #1}\else
  \providecommand{\doi}{doi: \begingroup \urlstyle{rm}\Url}\fi

\bibitem[Alon \& Yahav(2020)Alon and Yahav]{alon2020bottleneck}
Uri Alon and Eran Yahav.
\newblock On the bottleneck of graph neural networks and its practical
  implications.
\newblock \emph{arXiv preprint arXiv:2006.05205}, 2020.

\bibitem[Benczur \& Karger(2002)Benczur and Karger]{benczur2002randomized}
Andras Benczur and David~R Karger.
\newblock Randomized approximation schemes for cuts and flows in capacitated
  graphs.
\newblock \emph{arXiv preprint cs/0207078}, 2002.

\bibitem[Bhagat et~al.(2011)Bhagat, Cormode, and Muthukrishnan]{bhagat2011node}
Smriti Bhagat, Graham Cormode, and S~Muthukrishnan.
\newblock Node classification in social networks.
\newblock In \emph{Social network data analytics}, pp.\  115--148. Springer,
  2011.

\bibitem[Bojchevski \& G{\"u}nnemann(2017)Bojchevski and
  G{\"u}nnemann]{bojchevski2017deep}
Aleksandar Bojchevski and Stephan G{\"u}nnemann.
\newblock Deep gaussian embedding of graphs: Unsupervised inductive learning
  via ranking.
\newblock \emph{arXiv preprint arXiv:1707.03815}, 2017.

\bibitem[Cai et~al.(2018)Cai, Zheng, and Chang]{cai2018comprehensive}
Hongyun Cai, Vincent~W Zheng, and Kevin Chen-Chuan Chang.
\newblock A comprehensive survey of graph embedding: Problems, techniques, and
  applications.
\newblock \emph{IEEE Transactions on Knowledge and Data Engineering},
  30\penalty0 (9):\penalty0 1616--1637, 2018.

\bibitem[Calandriello et~al.(2018)Calandriello, Lazaric, Koutis, and
  Valko]{calandriello2018improved}
Daniele Calandriello, Alessandro Lazaric, Ioannis Koutis, and Michal Valko.
\newblock Improved large-scale graph learning through ridge spectral
  sparsification.
\newblock In \emph{International Conference on Machine Learning}, pp.\
  688--697. PMLR, 2018.

\bibitem[Chen et~al.(2020)Chen, Xu, Huang, Deng, Huang, Wang, He, and
  Li]{chen2020label}
Hao Chen, Yue Xu, Feiran Huang, Zengde Deng, Wenbing Huang, Senzhang Wang, Peng
  He, and Zhoujun Li.
\newblock Label-aware graph convolutional networks.
\newblock In \emph{Proceedings of the 29th ACM International Conference on
  Information \& Knowledge Management}, pp.\  1977--1980, 2020.

\bibitem[Chen et~al.(2018)Chen, Ma, and Xiao]{chen2018fastgcn}
Jie Chen, Tengfei Ma, and Cao Xiao.
\newblock Fastgcn: fast learning with graph convolutional networks via
  importance sampling.
\newblock \emph{arXiv preprint arXiv:1801.10247}, 2018.

\bibitem[Chu et~al.(2020)Chu, Gao, Peng, Sachdeva, Sawlani, and
  Wang]{chu2020graph}
Timothy Chu, Yu~Gao, Richard Peng, Sushant Sachdeva, Saurabh Sawlani, and
  Junxing Wang.
\newblock Graph sparsification, spectral sketches, and faster resistance
  computation via short cycle decompositions.
\newblock \emph{SIAM Journal on Computing}, pp.\  FOCS18--85, 2020.

\bibitem[Corso et~al.(2020)Corso, Cavalleri, Beaini, Li{\`o}, and
  Veli{\v{c}}kovi{\'c}]{corso2020principal}
Gabriele Corso, Luca Cavalleri, Dominique Beaini, Pietro Li{\`o}, and Petar
  Veli{\v{c}}kovi{\'c}.
\newblock Principal neighbourhood aggregation for graph nets.
\newblock \emph{arXiv preprint arXiv:2004.05718}, 2020.

\bibitem[Dwivedi et~al.(2020)Dwivedi, Joshi, Laurent, Bengio, and
  Bresson]{dwivedi2020benchmarking}
Vijay~Prakash Dwivedi, Chaitanya~K Joshi, Thomas Laurent, Yoshua Bengio, and
  Xavier Bresson.
\newblock Benchmarking graph neural networks.
\newblock \emph{arXiv preprint arXiv:2003.00982}, 2020.

\bibitem[Faber et~al.(2021)Faber, Lu, and Wattenhofer]{faber2021should}
Lukas Faber, Yifan Lu, and Roger Wattenhofer.
\newblock Should graph neural networks use features, edges, or both?
\newblock \emph{arXiv preprint arXiv:2103.06857}, 2021.

\bibitem[Fey \& Lenssen(2019)Fey and Lenssen]{fey2019fast}
Matthias Fey and Jan~Eric Lenssen.
\newblock Fast graph representation learning with pytorch geometric.
\newblock \emph{arXiv preprint arXiv:1903.02428}, 2019.

\bibitem[G{\'o}mez-Bombarelli et~al.(2018)G{\'o}mez-Bombarelli, Wei, Duvenaud,
  Hern{\'a}ndez-Lobato, S{\'a}nchez-Lengeling, Sheberla, Aguilera-Iparraguirre,
  Hirzel, Adams, and Aspuru-Guzik]{gomez2018automatic}
Rafael G{\'o}mez-Bombarelli, Jennifer~N Wei, David Duvenaud, Jos{\'e}~Miguel
  Hern{\'a}ndez-Lobato, Benjam{\'\i}n S{\'a}nchez-Lengeling, Dennis Sheberla,
  Jorge Aguilera-Iparraguirre, Timothy~D Hirzel, Ryan~P Adams, and Al{\'a}n
  Aspuru-Guzik.
\newblock Automatic chemical design using a data-driven continuous
  representation of molecules.
\newblock \emph{ACS central science}, 4\penalty0 (2):\penalty0 268--276, 2018.

\bibitem[Hamilton et~al.(2017)Hamilton, Ying, and
  Leskovec]{hamilton2017inductive}
William~L Hamilton, Rex Ying, and Jure Leskovec.
\newblock Inductive representation learning on large graphs.
\newblock In \emph{Proceedings of the 31st International Conference on Neural
  Information Processing Systems}, pp.\  1025--1035, 2017.

\bibitem[Hasanzadeh et~al.(2020)Hasanzadeh, Hajiramezanali, Boluki, Zhou,
  Duffield, Narayanan, and Qian]{hasanzadeh2020bayesian}
Arman Hasanzadeh, Ehsan Hajiramezanali, Shahin Boluki, Mingyuan Zhou, Nick
  Duffield, Krishna Narayanan, and Xiaoning Qian.
\newblock Bayesian graph neural networks with adaptive connection sampling.
\newblock In \emph{International conference on machine learning}, pp.\
  4094--4104. PMLR, 2020.

\bibitem[Hu \& Lau(2013)Hu and Lau]{hu2013survey}
Pili Hu and Wing~Cheong Lau.
\newblock A survey and taxonomy of graph sampling.
\newblock \emph{arXiv preprint arXiv:1308.5865}, 2013.

\bibitem[Hübler et~al.(2008)Hübler, Kriegel, Borgwardt, and
  Ghahramani]{4781123}
Christian Hübler, Hans-Peter Kriegel, Karsten Borgwardt, and Zoubin
  Ghahramani.
\newblock Metropolis algorithms for representative subgraph sampling.
\newblock In \emph{2008 Eighth IEEE International Conference on Data Mining},
  pp.\  283--292, 2008.
\newblock \doi{10.1109/ICDM.2008.124}.

\bibitem[Kim \& Oh(2020)Kim and Oh]{kim2020find}
Dongkwan Kim and Alice Oh.
\newblock How to find your friendly neighborhood: Graph attention design with
  self-supervision.
\newblock In \emph{International Conference on Learning Representations}, 2020.

\bibitem[Kipf \& Welling(2016)Kipf and Welling]{kipf2016semi}
Thomas~N Kipf and Max Welling.
\newblock Semi-supervised classification with graph convolutional networks.
\newblock \emph{arXiv preprint arXiv:1609.02907}, 2016.

\bibitem[Kitaev et~al.(2020)Kitaev, Kaiser, and Levskaya]{kitaev2020reformer}
Nikita Kitaev, {\L}ukasz Kaiser, and Anselm Levskaya.
\newblock Reformer: The efficient transformer.
\newblock \emph{arXiv preprint arXiv:2001.04451}, 2020.

\bibitem[Klicpera et~al.(2020)Klicpera, Gro{\ss}, and
  G{\"u}nnemann]{klicpera2020directional}
Johannes Klicpera, Janek Gro{\ss}, and Stephan G{\"u}nnemann.
\newblock Directional message passing for molecular graphs.
\newblock \emph{arXiv preprint arXiv:2003.03123}, 2020.

\bibitem[Kriege et~al.(2020)Kriege, Johansson, and Morris]{kriege2020survey}
Nils~M Kriege, Fredrik~D Johansson, and Christopher Morris.
\newblock A survey on graph kernels.
\newblock \emph{Applied Network Science}, 5\penalty0 (1):\penalty0 1--42, 2020.

\bibitem[Kumar et~al.(2020)Kumar, Singh, Singh, and Biswas]{kumar2020link}
Ajay Kumar, Shashank~Sheshar Singh, Kuldeep Singh, and Bhaskar Biswas.
\newblock Link prediction techniques, applications, and performance: A survey.
\newblock \emph{Physica A: Statistical Mechanics and its Applications},
  553:\penalty0 124289, 2020.

\bibitem[Li et~al.(2018)Li, Han, and Wu]{li2018deeper}
Qimai Li, Zhichao Han, and Xiao-Ming Wu.
\newblock Deeper insights into graph convolutional networks for semi-supervised
  learning.
\newblock In \emph{Thirty-Second AAAI conference on artificial intelligence},
  2018.

\bibitem[Luo et~al.(2021)Luo, Cheng, Yu, Zong, Ni, Chen, and
  Zhang]{luo2021learning}
Dongsheng Luo, Wei Cheng, Wenchao Yu, Bo~Zong, Jingchao Ni, Haifeng Chen, and
  Xiang Zhang.
\newblock Learning to drop: Robust graph neural network via topological
  denoising.
\newblock In \emph{Proceedings of the 14th ACM International Conference on Web
  Search and Data Mining}, pp.\  779--787, 2021.

\bibitem[Paszke et~al.(2019)Paszke, Gross, Massa, Lerer, Bradbury, Chanan,
  Killeen, Lin, Gimelshein, Antiga, et~al.]{paszke2019pytorch}
Adam Paszke, Sam Gross, Francisco Massa, Adam Lerer, James Bradbury, Gregory
  Chanan, Trevor Killeen, Zeming Lin, Natalia Gimelshein, Luca Antiga, et~al.
\newblock Pytorch: An imperative style, high-performance deep learning library.
\newblock \emph{Advances in neural information processing systems},
  32:\penalty0 8026--8037, 2019.

\bibitem[Paulev{\'e} et~al.(2010)Paulev{\'e}, J{\'e}gou, and
  Amsaleg]{pauleve2010locality}
Lo{\"\i}c Paulev{\'e}, Herv{\'e} J{\'e}gou, and Laurent Amsaleg.
\newblock Locality sensitive hashing: A comparison of hash function types and
  querying mechanisms.
\newblock \emph{Pattern recognition letters}, 31\penalty0 (11):\penalty0
  1348--1358, 2010.

\bibitem[Rivest(1992)]{rivest1992rfc1321}
Ronald Rivest.
\newblock Rfc1321: The md5 message-digest algorithm, 1992.

\bibitem[Rong et~al.(2019)Rong, Huang, Xu, and Huang]{rong2019dropedge}
Yu~Rong, Wenbing Huang, Tingyang Xu, and Junzhou Huang.
\newblock Dropedge: Towards deep graph convolutional networks on node
  classification.
\newblock \emph{arXiv preprint arXiv:1907.10903}, 2019.

\bibitem[Rozemberczki et~al.(2019)Rozemberczki, Allen, and
  Sarkar]{rozemberczki2019multi}
Benedek Rozemberczki, Carl Allen, and Rik Sarkar.
\newblock Multi-scale attributed node embedding.(2019).
\newblock \emph{arXiv preprint cs.LG/1909.13021}, 2019.

\bibitem[Rozemberczki et~al.(2021)Rozemberczki, Allen, and
  Sarkar]{rozemberczki2021multi}
Benedek Rozemberczki, Carl Allen, and Rik Sarkar.
\newblock Multi-scale attributed node embedding.
\newblock \emph{Journal of Complex Networks}, 9\penalty0 (2):\penalty0 cnab014,
  2021.

\bibitem[Sadhanala et~al.(2016)Sadhanala, Wang, and
  Tibshirani]{sadhanala2016graph}
Veeru Sadhanala, Yu-Xiang Wang, and Ryan Tibshirani.
\newblock Graph sparsification approaches for laplacian smoothing.
\newblock In \emph{Artificial Intelligence and Statistics}, pp.\  1250--1259.
  PMLR, 2016.

\bibitem[Shakhnarovich et~al.(2008)Shakhnarovich, Darrell, and
  Indyk]{shakhnarovich2008nearest}
Gregory Shakhnarovich, Trevor Darrell, and Piotr Indyk.
\newblock Nearest-neighbor methods in learning and vision.
\newblock \emph{IEEE Trans. Neural Networks}, 19\penalty0 (2):\penalty0 377,
  2008.

\bibitem[Spielman \& Srivastava(2008)Spielman and
  Srivastava]{spielman2008graph}
DA~Spielman and N~Srivastava.
\newblock Graph sparsification by effective resistances. corr abs/0803.0929
  (2008), 2008.

\bibitem[Spielman \& Teng(2008)Spielman and Teng]{spielman2008spectral}
Daniel~A Spielman and Shang-Hua Teng.
\newblock Spectral sparsification of graphs. corr, abs/0808.4134, 2008.
\newblock \emph{arXiv preprint arXiv:0808.4134}, 2008.

\bibitem[Srinivasa et~al.(2020)Srinivasa, Xiao, Glass, Romberg, and
  Sun]{srinivasa2020fast}
Rakshith~S Srinivasa, Cao Xiao, Lucas Glass, Justin Romberg, and Jimeng Sun.
\newblock Fast graph attention networks using effective resistance based graph
  sparsification.
\newblock \emph{arXiv preprint arXiv:2006.08796}, 2020.

\bibitem[Sterling \& Irwin(2015)Sterling and Irwin]{sterling2015zinc}
Teague Sterling and John~J Irwin.
\newblock Zinc 15--ligand discovery for everyone.
\newblock \emph{Journal of chemical information and modeling}, 55\penalty0
  (11):\penalty0 2324--2337, 2015.

\bibitem[Vaswani et~al.(2017)Vaswani, Shazeer, Parmar, Uszkoreit, Jones, Gomez,
  Kaiser, and Polosukhin]{vaswani2017attention}
Ashish Vaswani, Noam Shazeer, Niki Parmar, Jakob Uszkoreit, Llion Jones,
  Aidan~N Gomez, {\L}ukasz Kaiser, and Illia Polosukhin.
\newblock Attention is all you need.
\newblock In \emph{Advances in neural information processing systems}, pp.\
  5998--6008, 2017.

\bibitem[Veli{\v{c}}kovi{\'c} et~al.(2017)Veli{\v{c}}kovi{\'c}, Cucurull,
  Casanova, Romero, Lio, and Bengio]{velivckovic2017graph}
Petar Veli{\v{c}}kovi{\'c}, Guillem Cucurull, Arantxa Casanova, Adriana Romero,
  Pietro Lio, and Yoshua Bengio.
\newblock Graph attention networks.
\newblock \emph{arXiv preprint arXiv:1710.10903}, 2017.

\bibitem[Wu et~al.(2017)Wu, Ramsundar, Feinberg, Gomes, Geniesse, Pappu,
  Leswing, and Pande]{wu2017moleculenet}
Z~Wu, B~Ramsundar, EN~Feinberg, J~Gomes, C~Geniesse, AS~Pappu, K~Leswing, and
  V~Pande.
\newblock Moleculenet: a benchmark for molecular machine learning. arxiv
  e-prints.
\newblock \emph{arXiv preprint arXiv:1703.00564}, 2017.

\bibitem[Wu et~al.(2020)Wu, Pan, Chen, Long, Zhang, and
  Philip]{wu2020comprehensive}
Zonghan Wu, Shirui Pan, Fengwen Chen, Guodong Long, Chengqi Zhang, and S~Yu
  Philip.
\newblock A comprehensive survey on graph neural networks.
\newblock \emph{IEEE transactions on neural networks and learning systems},
  32\penalty0 (1):\penalty0 4--24, 2020.

\bibitem[Xu et~al.(2018)Xu, Hu, Leskovec, and Jegelka]{xu2018powerful}
Keyulu Xu, Weihua Hu, Jure Leskovec, and Stefanie Jegelka.
\newblock How powerful are graph neural networks?
\newblock \emph{arXiv preprint arXiv:1810.00826}, 2018.

\bibitem[Ye \& Ji(2021)Ye and Ji]{ye2021sparse}
Yang Ye and Shihao Ji.
\newblock Sparse graph attention networks.
\newblock \emph{IEEE Transactions on Knowledge and Data Engineering}, 2021.

\bibitem[Zhang et~al.(2018)Zhang, Yin, Zhu, and Zhang]{zhang2018network}
Daokun Zhang, Jie Yin, Xingquan Zhu, and Chengqi Zhang.
\newblock Network representation learning: A survey.
\newblock \emph{IEEE transactions on Big Data}, 6\penalty0 (1):\penalty0 3--28,
  2018.

\bibitem[Zhang et~al.(2020)Zhang, Liu, and Xie]{zhang2020molecular}
Shuo Zhang, Yang Liu, and Lei Xie.
\newblock Molecular mechanics-driven graph neural network with multiplex graph
  for molecular structures.
\newblock \emph{arXiv preprint arXiv:2011.07457}, 2020.

\bibitem[Zheng et~al.(2020)Zheng, Zong, Cheng, Song, Ni, Yu, Chen, and
  Wang]{zheng2020robust}
Cheng Zheng, Bo~Zong, Wei Cheng, Dongjin Song, Jingchao Ni, Wenchao Yu, Haifeng
  Chen, and Wei Wang.
\newblock Robust graph representation learning via neural sparsification.
\newblock In \emph{International Conference on Machine Learning}, pp.\
  11458--11468. PMLR, 2020.

\bibitem[Zitnik \& Leskovec(2017)Zitnik and Leskovec]{zitnik2017predicting}
Marinka Zitnik and Jure Leskovec.
\newblock Predicting multicellular function through multi-layer tissue
  networks.
\newblock \emph{Bioinformatics}, 33\penalty0 (14):\penalty0 i190--i198, 2017.

\end{thebibliography}
\bibliographystyle{iclr2022_conference}

\appendix
\section{Construction of Edge Attributes}\label{app:edge_attrs}
As discussed in Section \ref{section:methodology}, LSP assumes the existence of edge attributes. However, many graph datasets do not include such representation but include node attributes instead (or additionally). Additionally, some datasets have scalar values as edge and node representations (for instance, molecules with number of bonds as edge attributes and atomic number as node attribute). Those representations cannot be used as inputs to our hash functions because they expect vectors as input rather than scalars. Here we describe our choice of representation in each case.

Recall the definition of edge attributes from Section \ref{section:methodology} - an attribute of an edge \text{$(u,v)\in~E$} is a $d$-dimensional vector denoted as \text{$e_{u,v}\in~\Real^{d}$}. Similarly, an attribute of a node \text{$u\in~V$} is a $q$-dimensional vector. For convenience, we use the name of a node itself as a notation for its attribute. We denote the final representation of an edge attribute which serves as an input to the $LSP$ algorithm as $e_{u,v}^{LSP}$.

\textbf{Node attributes only -} \text{$e_{u,v}^{LSP} = [u, v]$}.

\textbf{Both node and edge attributes -} \text{$e_{u,v}^{LSP} = [u, e_{u,v}, v]$}.

\textbf{Scalar node/edge attributes -}Let $r\in \Natural$ be an attribute of a node or edge (for example, atomic number or number of bonds between atoms in a graphs that represents a molecule). Assume $r$ accepts $m$ possible values, $r\in \{0,...,m-1\}$. We generate a random matrix $M\in \Real^{m\times m}$ and use the columns of $M$ as embeddings of the corresponding component; that is, the $r^{th}$ column of $M$ is the corresponding feature vector of the graph component associated with the attribute $r$.

\begin{example}
We finalize this section with a complete example of constructing edge attributes from a graph \text{$G=(V,E)$} representing a molecule. We assume every node $v \in V$ is associated with a natural scalar \text{$r_v\in \{0,...,m_V-1\}$} and every edge \text{$e\in E$} is associated with a natural scalar \text{$r_e\in \{0,...,m_E-1\}$}. Following our explanation for generating embeddings for scalar attributes, we generate two random real matrices, \text{$M_V \in \Real^{m_V\times m_V}$} and \text{$M_E\in \Real^{m_E\times m_E}$}. Then, the columns of each matrix are used as feature vectors of nodes and edges as follow. Let $M^i$ denote the $i^{th}$ column of a matrix $M$. Then, following our explanation for constructing edge representations as input for $LSP$, given two nodes \text{$u,v\in V$} and an edge connecting the two nodes $e_{u,v}$, we obtain the following:
\(
    e_{u,v}^{LSP} = [M_V^u,M_E^{e_{u,v}}, M_V^v]
\).
\end{example}

\section{Experiments}\label{ref:app_experiments}
\subsection{Details}\label{app:experiments_details}
\paragraph{Implementation Details -} All experiments were run on Ubuntu 18.04 machine with Intel i9-10920X CPU, 93GB of RAM and 2 GeForce RTX 3090 GPUs. Our implementation of LSP is in Python 3. To generate graphs and perform training of graph neural models, we use Pytorch-Geometric \citep{fey2019fast} and use Pytorch \citep{paszke2019pytorch}. Moreover, we all datasets used in this section are retrieved from the collection supplied by Pytorch-Geometric.

\subsection{Datasets}\label{app:datasets}
\subsubsection{Synthetic Graph Dataset Generator}\label{app:syn_generator}
Here, we introduce the synthetic graph classification dataset generator. Table \ref{table:syn_dataset_parameters} summarizes the parameters used for the synthetic data generation and their default values, which we use for our experiment described in Section \ref{section:graph_classification_experiments}.

\bgroup
\def\arraystretch{1.2}
\begin{table}[!h]
\centering
 \caption{Parameters of the synthetic dataset generator. For each parameter introduced in this table, the default parameter denotes the value that we use for our experiments in Section \ref{section:graph_classification_experiments}.} \label{table:syn_dataset_parameters}
\begin{tabular}{ |p{3.6cm}|p{8cm}|p{1.2cm}| }
 \hline
 \textbf{Parameter}&\textbf{Description}&\textbf{Default}\\ 
 \hline
 Number of samples&The number of graph instances in the dataset&20,000\\
 \hline
 Number of classes&The number of classes in the dataset&100\\
 \hline
 Minimum nodes&Minimum number of nodes in each graph sample&40\\
 \hline
 Maximum nodes&Maximum number of nodes in each graph sample&60\\
 \hline
 Node dimension&Dimensionality of the node representation vectors &10\\
 \hline
 Edge dimension&Dimensionality of the edge representation vectors&40\\
 \hline
 Connectivity rate&Degree of each node&0.2\\
 \hline
 Node centers std&Scatter of the node representation vectors&0.2\\
 \hline
 Edge centers std&Scatter of the edge representation vectors&0.2\\
 \hline
 Node noise std&Amount of additive noise for node representation vectors&0.25\\
 \hline
 Edge noise std&Amount of additive noise for edge representation vectors&0.1\\
 \hline
 Is symmetric&Whether to enforce symmetry on each graph&False\\
 \hline
 Node removal probability&The probability of remove a node from a graph sample, after the graph is initally generated&0.1\\
 \hline
\end{tabular}
\end{table}
\egroup

\subsubsection{Real-World datasets}\label{app:real_world_datasets}
Here, we describe the various real-world datasets used in this paper for different tasks.

\paragraph{Node Classification}
\begin{itemize}
    \item \textbf{GitHub} - A social network where nodes correspond to developers who have starred at least 10
repositories and edges to mutual follower relationships. Node features are location, starred repositories, employer and e-mail address. The task is to classify nodes as web or machine learning
developers.
    
    
    
    \item \textbf{Cora, CiteSeer and PubMed} - These are citation network with labels of paper topic. Each node represents a publication corresponding to different class (seven classes for Cora, 6 classes for CiteSeer and 3 classes for PubMed). The citation network consists of links representing citations. In Cora and CiteSeer, each publication in the dataset is described by a 0/1-valued word vector indicating the absence/presence of the corresponding word from the dictionary. In PubMed, each publication in the dataset is described by a TF/IDF weighted word vector from a dictionary which consists of 500 unique words.

    \item \textbf{PPI} - This dataset is made up of 24 graphs, with each graph corresponding to a different human tissue. Each node represents a protein and is associated with features made up of motif gene sets and immunological signatures. The task is to classify each protein into gene ontology sets  (121 in total).
\end{itemize}

Statistics for these datasets are shown in Table \ref{table:data_stats_node_classification}. 

\bgroup
\def\arraystretch{1.2}
\begin{table}[!h]
\caption{Node Classification Datasets Statistics} 
\label{table:data_stats_node_classification}
\centering
\begin{tabular}{ |c|c|c|c|c|c|c| }
\hline
Dataset & Github & Cora & CiteSeer & PubMed & PPI\\
\hline
\hline
Number of nodes & 37700 & 2708 & 3327 & 19717 & 44906 \\
Number of edges & 578006 & 10556 & 9104 & 88648 & 1136460 \\
Number of classes & 2 & 7 & 6 & 3 & 121 \\
Average degree & 30 & 7 & 5 & 8 & 50 \\
\hline
\end{tabular}
\end{table}
\egroup

\paragraph{Graph Regression}
\begin{itemize}
\item \textbf{QM9} - The Quantum Mechanics 9 database contains around 130k small organic molecules with up to 9 heavy atoms and their physical properties in equilibrium, computed using density functional theory calculations. Following previous works \citep{zhang2020molecular, klicpera2020directional}, we randomly sample 110k molecules for training, 10k for validation and the rest for testing. Results are presented in terms of Mean Absolute Error (MAE).
\item \textbf{ZINC} - A collection of chemical compounds prepared especially for virtual screening. In this collection, a graph represents a molecule, with node
features indicating the atom type and edge features the type of chemical bond between two atoms. The goal is to regress the molecule's penalised water-octanol partition coefficient. We use the original train/validation/test splits from \cite{gomez2018automatic} as provided in Pytorch-Geometric \citep{fey2019fast}. In similar to \cite{dwivedi2020benchmarking}, we randomly select 12K for efficiency.
\end{itemize}


\subsection{Models}\label{app:models}
As stated in Section \ref{section:experiments}, we use the state-of-the-art model identified with each dataset to compare performance of LSP with other baselines. In the following, we provide descriptions of these models. After introducing these models, we present the basic architectures used as baselines for our experiments.

\subsubsection{Node classification}
\begin{itemize}
    \item \textbf{GitHub} - GAT model used in the comparisons in \citep{rozemberczki2021multi}. This model consists of 2 attention layers that aggregate information up to 2-hop neighbourhoods with 1 attention head. For non-linearities we used leaky rectified linear unit (Leaky ReLU) with negative slope of 0.2.
    
    
    
    
    \item \textbf{Cora}, \textbf{CiteSeer} and \textbf{PubMed} - We use the methodology introduced in GraphSage \citep{hamilton2017inductive} in combination with GAT, which has 2 layers. Following GraphSage, we sample at most 10, 15 neighbors for the first and second hop respectively for training. For testing, we sample the whole neighborhood of a node. The first layer has 2 attention heads, where each attention result is has hidden size of $8$ and concatenated to the others. The second layer, which is used for calculating the output has 1 attention head. We use exponential linear unit (ELU) as activation between the layers.

    \item \textbf{PPI} - GAT model with 3 layers from \cite{vaswani2017attention}. The hidden dimension of each layer is 256 and the number of attention heads in each layer is 4,4,6 for the first, second and third layer respectively. For non-linearities we used exponential linear unit (ELU).
\end{itemize}

\subsubsection{Graph regression}
\begin{itemize}
    \item \textbf{QM9} - For all experiments, we use the MXMNet architecture proposed in \citep{zhang2020molecular}, with best performing configurations presented in the original paper.

    \item \textbf{ZINC} - We use the PNA architecture \citep{corso2020principal} with its default settings.
\end{itemize}

\subsubsection{Graph classification}
\begin{itemize}
    \item \textbf{Synthetic dataset} - We construct a GAT model with 3 layers with 16 attention heads each. The hidden size of the intermediate layers is 40. The first two layers concatenate the outputs of the attention heads for output vector of size \text{$40\cdot 16=640$}. For non-linearities, we apply ReLU between every two consecutive layers. Then, in order to perform the final classification, we average the the result of the third attention layer and aggregate the obtained node representations using global mean pooling. The result is later fed into a dropout layer which cancels 50\% of the activations. Finally, we feed the result into a linear layer which provides the final prediction of the model.
\end{itemize}

\subsubsection{Basic models}\label{app:basic_models}
The basic models used as baselines demonstrate the time-performance trade-off via training computationally cheaper models that deliver less performance when compared to computationally extensive models trained on pruned graphs. As we use GATs \citep{velivckovic2017graph} in all experiments of node-classification and graph classification, the basic models for these experiments are created by removing the attention mechanism while keeping all the accompanied configurations as is. As a result, we obtain GCN \citep{kipf2016semi} variants variant of the original models that aggregate messages from neighboring nodes by averaging instead if weights computed via attention heads.

\section{Reducing the Neighborhoods Variance}\label{app:neighborhood_variance}

The depth of a node's neighborhood determines the maximal path length between a node and other nodes contained in its neighborhood. A neighborhood of depth $k$ is called a k-hop neighborhood. As we increase the depth of the neighborhoods, the number of nodes contained within the neighborhoods increases. However, not all neighborhoods increase in the same way, which in turn leads to a variable size of neighborhoods - a phenomenon we call \textit{Neighborhood Variance}, as exemplified in Figure \ref{fig:neighborhood_variance}. For the two given nodes $v_1,v_2$, their 3-hop neighborhoods have different sizes, which in turn translates to a variable amount of information that has to be encoded within a fixed-length code. 

To see this, we measure the amount of nodes reachable within $k$ hops for each node $v$ in a graph $G$, denoted as $\neighborhood_{G}^{k,v}$. We claim the following:
\begin{proposition}[Pruning edges for reducing prevents over-squashing]\label{proposition:over_squashing_prevention}
    Let \text{$G=(V,E)$} be a graph and consider a sparsified version \text{$G'=(V,E')$} so that \text{$E'\subset E$}. For each \text{$v\in V$}, denote by \text{$\neighborhood_{G}^{k,v}$} the set of nodes reachable from $v$ within $k$ hops in the graph $G$. Then, there exists \text{$k_{max}\in \mathbb{N}$} so that for all \text{$k\leq k_{max}$} the variance of the amount of neighboring nodes needed to be encoded into the representation vector of a node satisfies:
    \begin{equation}
        Var(|\neighborhood_{G'}^{v,k}|) \leq Var(|\neighborhood_{G}^{v,k}|).
    \end{equation}
\end{proposition}
Suppose that a $k$-layer GCN (for \text{$k \leq k_{max}$}) is trained to solve a task by aggregating information from $k$-hop neighborhoods into node representations. Then, from Proposition \ref{proposition:over_squashing_prevention} we know that the neighborhood size of each node in the pruned graph has less variance and thus a similar amount of information has to be encoded within the representation vector of each node. Consequently, over-squashing is avoided. To demonstrate the scope of this phenomenon, we provide statistics about neighborhood size variance from popular real-world datasets in Appendix \ref{app:var_statistics}. Additionally, we justify Proposition \ref{proposition:over_squashing_prevention} in Appendix \ref{proposition:over_squashing_prevention}.

\subsection{Neighborhood variance in Real-World graphs }\label{app:var_statistics}
\begin{figure}[!ht]
    \centering
    \begin{subfigure}[b]{0.32\textwidth}
         \centering
         \begin{tikzpicture}
    \begin{axis}[
        height=1.3in,
        width=\linewidth,
        ymode=log,
        log basis y={2},
        title={CiteSeer},
        xlabel={\% Edges ($\cdot 100$)},
        ylabel={$Var(|\neighborhood_{G'}^{v,k}|)$},
        xmin=0, xmax=1,
        xtick={0,0.2,0.4,0.6,0.8,1},
        xticklabel style={rotate=0},
        ymajorgrids=true,
        xmajorgrids=true,
        grid style=dashed,
        enlargelimits=0.1,
        legend pos=south west,
        label style={font=\footnotesize},
        xticklabel style={font=\scriptsize},
        yticklabel style={font=\scriptsize},
    ]
     
    \addplot[
        color=blue,mark=square,style=solid,line width=1pt
        ]
        coordinates {
(0.1, 0.3459864996875488)
(0.2, 0.954730489282206)
(0.30000000000000004, 1.6444942414934633)
(0.4, 2.232751696211982)
(0.5, 3.6705607200118457)
(0.6, 4.647108315538025)
(0.7000000000000001, 6.013254398867316)
(0.8, 7.44577149243617)
(0.9, 9.31682929757703)
(1.0, 11.430063739680685)
        };
        
        \addplot[
        color=purple,mark=pentagon,style=solid,line width=1pt
        ]
        coordinates {
(0.1, 3.741364137397575)
(0.2, 33.0147189488703)
(0.30000000000000004, 158.44896068987342)
(0.4, 399.14139299294453)
(0.5, 760.6311181506359)
(0.6, 1204.915380340772)
(0.7000000000000001, 1889.458155707747)
(0.8, 2550.455904993157)
(0.9, 3838.8838092646542)
(1.0, 4982.347807272049)
        };
        
        \addplot[
        color=teal,mark=triangle,style=solid,line width=1pt
        ]
        coordinates {
(0.1, 5.595737762885642)
(0.2, 211.07664689149237)
(0.30000000000000004, 1083.8742048124077)
(0.4, 3745.8030060541532)
(0.5, 7037.8703860147625)
(0.6, 13997.515381479092)
(0.7000000000000001, 21633.7015713083)
(0.8, 29475.57897082907)
(0.9, 39978.288562515845)
(1.0, 49957.7975484349)
        };
        
        \addplot[
        color=gray,mark=star,style=solid,line width=1pt
        ]
        coordinates {
(0.1, 16.00192340198406)
(0.2, 490.67308463176516)
(0.30000000000000004, 3012.831760868644)
(0.4, 9875.804444675721)
(0.5, 22841.274795601268)
(0.6, 40869.98889919702)
(0.7000000000000001, 68191.04545399109)
(0.8, 77166.73044573689)
(0.9, 134768.27914697077)
(1.0, 179792.1669961023)
        };
        
        \addplot[
        color=black,mark=diamond,style=solid,line width=1pt
        ]
        coordinates {
(0.1, 44.63846719045718)
(0.2, 699.7132351287102)
(0.30000000000000004, 6671.3829919769105)
(0.4, 15015.047720877055)
(0.5, 37651.49716128814)
(0.6, 74381.28408304001)
(0.7000000000000001, 131214.61458755404)
(0.8, 236519.4763836682)
(0.9, 349701.10461960686)
(1.0, 443442.4469662783)
        };

    \end{axis}
    \end{tikzpicture}
     \end{subfigure}
    \begin{subfigure}[b]{0.32\textwidth}
         \centering
         \begin{tikzpicture}
    \begin{axis}[
        height=1.3in,
        width=\linewidth,
        ymode=log,
        log basis y={2},
        title={Cora},
        xlabel={\% Edges ($\cdot 100$)},
        ylabel={$Var(|\neighborhood_{G'}^{v,k}|)$},
        xmin=0, xmax=1,
        xtick={0,0.2,0.4,0.6,0.8,1},
        xticklabel style={rotate=0},
        ymajorgrids=true,
        xmajorgrids=true,
        grid style=dashed,
        enlargelimits=0.1,
        legend pos=south west,
        label style={font=\footnotesize},
        xticklabel style={font=\scriptsize},
        yticklabel style={font=\scriptsize},
    ]
     
    \addplot[
        color=blue,mark=square,style=solid,line width=1pt
        ]
        coordinates {
(0.1, 0.5967453237739704)
(0.2, 1.6925356839737393)
(0.30000000000000004, 3.1006422242537566)
(0.4, 5.428192957460689)
(0.5, 7.44053452869009)
(0.6, 10.905642426073843)
(0.7000000000000001, 13.326996955243942)
(0.8, 18.644825005618237)
(0.9, 22.73664005550598)
(1.0, 27.330084938984882)
        };
        
        \addplot[
        color=purple,mark=pentagon,style=solid,line width=1pt
        ]
        coordinates {
(0.1, 7.021420202518279)
(0.2, 75.84857765927968)
(0.30000000000000004, 316.11501058737286)
(0.4, 1027.9970843815252)
(0.5, 2609.2339754848595)
(0.6, 4652.263459627254)
(0.7000000000000001, 6945.126064873703)
(0.8, 9947.182132812892)
(0.9, 13773.563430826984)
(1.0, 18103.46521439839)
        };
        
        \addplot[
        color=teal,mark=triangle,style=solid,line width=1pt
        ]
        coordinates {
(0.1, 7.346801506123329)
(0.2, 584.0730494633767)
(0.30000000000000004, 5230.384446952953)
(0.4, 12059.45673454549)
(0.5, 35647.23008512989)
(0.6, 62516.59043108225)
(0.7000000000000001, 132424.53071960862)
(0.8, 185787.3463587292)
(0.9, 252926.21214127843)
(1.0, 311630.4444798387)
        };
        
        \addplot[
        color=gray,mark=star,style=solid,line width=1pt
        ]
        coordinates {
(0.1, 42.65322453957746)
(0.2, 688.3599499486178)
(0.30000000000000004, 16713.450547532448)
(0.4, 65639.5904834464)
(0.5, 173467.0782936493)
(0.6, 301864.87996013777)
(0.7000000000000001, 471633.1254404587)
(0.8, 554738.727719062)
(0.9, 576479.3869146127)
(1.0, 542467.0113815622)
        };
        
        \addplot[
        color=black,mark=diamond,style=solid,line width=1pt
        ]
        coordinates {
(0.1, 20.61680187703593)
(0.2, 1376.2441821268128)
(0.30000000000000004, 26808.601549732837)
(0.4, 207073.78518092356)
(0.5, 435396.79861355055)
(0.6, 587693.688520146)
(0.7000000000000001, 688810.9253477033)
(0.8, 645386.3286841984)
(0.9, 565085.0389396864)
(1.0, 496845.8606612281)

        };

    \end{axis}
    \end{tikzpicture}
     \end{subfigure}
    \begin{subfigure}[b]{0.32\textwidth}
         \centering
         \begin{tikzpicture}
    \begin{axis}[
        height=1.3in,
        width=\linewidth,
        ymode=log,
        log basis y={2},
        title={PubMed},
        xlabel={\% Edges ($\cdot 100$)},
        ylabel={$Var(|\neighborhood_{G'}^{v,k}|)$},
        xmin=0, xmax=1,
        xtick={0,0.2,0.4,0.6,0.8,1},
        xticklabel style={rotate=0},
        ymajorgrids=true,
        xmajorgrids=true,
        grid style=dashed,
        enlargelimits=0.1,
        legend pos=south west,
        label style={font=\footnotesize},
        xticklabel style={font=\scriptsize},
        yticklabel style={font=\scriptsize},
    ]
     
    \addplot[
        color=blue,mark=square,style=solid,line width=1pt
        ]
        coordinates {
(0.1, 0.9554743979904787)
(0.2, 2.975278010084003)
(0.30000000000000004, 5.842288553442533)
(0.4, 9.876838735881655)
(0.5, 15.092115147653443)
(0.6, 20.972237847182917)
(0.7000000000000001, 27.994262615779988)
(0.8, 36.013314267967466)
(0.9, 45.13447978452335)
(1.0, 55.21980714434912)
        };
        
        \addplot[
        color=purple,mark=pentagon,style=solid,line width=1pt
        ]
        coordinates {
(0.1, 39.00736239413712)
(0.2, 689.0764300601855)
(0.30000000000000004, 3948.683448315601)
(0.4, 12271.739144241217)
(0.5, 30873.7535512088)
(0.6, 60568.268997679945)
(0.7000000000000001, 96489.24142748614)
(0.8, 156984.39506239543)
(0.9, 221643.8401688863)
(1.0, 307770.5499249282)
        };
        
        \addplot[
        color=teal,mark=triangle,style=solid,line width=1pt
        ]
        coordinates {
(0.1, 545.6328502949798)
(0.2, 31130.656983325258)
(0.30000000000000004, 242084.90008197833)
(0.4, 785607.490417989)
(0.5, 1921276.492277385)
(0.6, 3278635.385363825)
(0.7000000000000001, 5899301.439558525)
(0.8, 8018989.212099198)
(0.9, 10712055.14255353)
(1.0, 12858301.442504294)
        };
        
        \addplot[
        color=gray,mark=star,style=solid,line width=1pt
        ]
        coordinates {
(0.1, 4089.871491365977)
(0.2, 276622.42388657853)
(0.30000000000000004, 2085288.010431899)
(0.4, 7020581.394668989)
(0.5, 14188827.78948175)
(0.6, 21715396.8497045)
(0.7000000000000001, 26996333.642668553)
(0.8, 28716121.21414679)
(0.9, 23731910.529562708)
(1.0, 14750268.028029827)

        };
        
        \addplot[
        color=black,mark=diamond,style=solid,line width=1pt
        ]
        coordinates {
(0.1, 16589.158201741226)
(0.2, 1140338.4361236575)
(0.30000000000000004, 7830613.6936642565)
(0.4, 19282090.242164273)
(0.5, 30911646.040004518)
(0.6, 36425709.22677841)
(0.7000000000000001, 36082394.48232284)
(0.8, 30470885.28802155)
(0.9, 18021809.15992668)
(1.0, 1722951.1849030831)

        };

    \end{axis}
    \end{tikzpicture}
     \end{subfigure}
     
     
     \begin{subfigure}[b]{\textwidth}
         \centering
         \begin{tikzpicture}
    \begin{axis}[
        axis lines=none,
        height=0.8in,
        width=0.8in,
        title={},
        xmin=0, xmax=1,
        ymin=0, ymax=1,
        ytick={},
        xtick={},
        legend columns=5,
        legend style={at={(0.5,0)},anchor=west}
    ]
     \addplot+[forget plot]
       coordinates{
   (0, 2.388854806)
       };
    \addlegendimage{color=blue,mark=square,style=solid,line width=1pt}
    \addlegendentry{k=1}
    \addlegendimage{color=purple,mark=pentagon,style=solid,line width=1pt}
    \addlegendentry{k=3}
    \addlegendimage{color=teal,mark=triangle,style=solid,line width=1pt}
    \addlegendentry{k=5}
    \addlegendimage{color=gray,mark=star,style=solid,line width=1pt}
    \addlegendentry{k=7}
    \addlegendimage{color=black,mark=diamond,style=solid,line width=1pt}
    \addlegendentry{k=9}
    \end{axis}
    \end{tikzpicture}
     \end{subfigure}
    \caption{Variance of neighborhood sizes as a function of edge percentage preserved from the original graph (Log scale).}
    \label{fig:neighborhood_variance}
\end{figure}
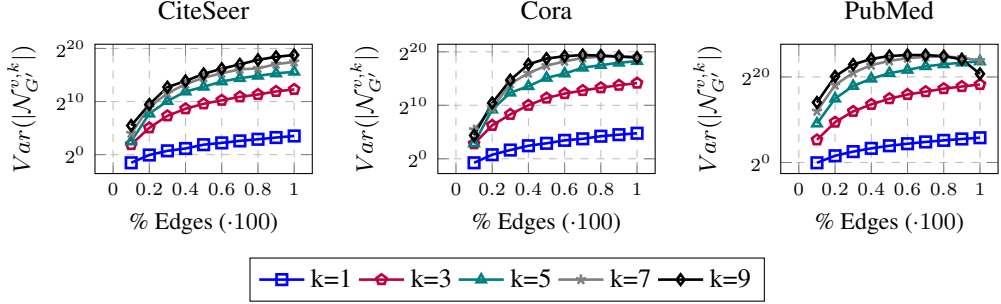

Figure \ref{fig:neighborhood_variance} depicts this variance for various neighborhood depths and pruning settings for certain datasets. For all datasets, we observe an exponential upward trend in the variance of the neighborhood sizes as more edges are incorporated in the graph for certain values of neighborhood depths ($k$). Additionally, an increase in the depth of the neighborhood (higher $k$ values) leads to a significant increase in this variance, which explains the performance degradation of deep GCNs due to over-squashing. This verifies our conclusion from Proposition \ref{proposition:over_squashing_prevention} for real-world scenarios.

\subsection{Justification of Proposition \ref{proposition:over_squashing_prevention}}
This claim is true for \text{$k_{max}=1$}, thus proving its existence:

The size of the $1$-hop neighborhood of $v$ equals the number of edges connected to it. Assuming the ratio of preserved edges in the pruned graph $G'$ is \text{$0\leq p<1$}, we have:
\begin{align*}
    Var(|\neighborhood_{G'}^{v,1}|) = 
    Var(d_{G'}(v))) = 
    E[d_{G'}(v)^2] - E[d_{G'}(v)]^2 = \\
    = E[(p\cdot d_{G}(v))^2] - E[p\cdot d_{G}(v)]^2 =
    E[p^2\cdot d_{G}(v)^2] - E[p\cdot d_{G}(v)]^2 = \\
    = p^2\cdot E[ d_{G}(v)^2] - p^2\cdot E[d_{G}(v)]^2 = 
    p^2\cdot ( E[ d_{G}(v)^2] - E[d_{G}(v)]^2) = \\
    = p^2\cdot Var(d_{G}(v))) \leq Var(d_{G}(v))) = Var(|\neighborhood_{G}^{v,1}|)
\end{align*}

\subsection{Additional Results}\label{app:additional_results}
\subsection{Node Classification}
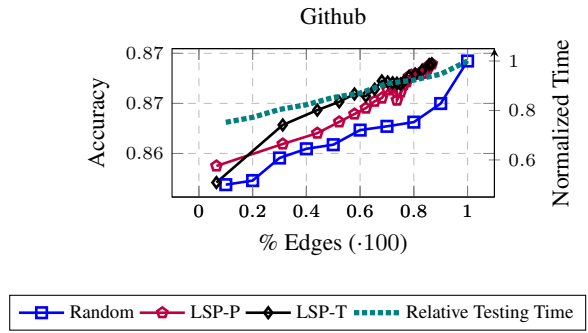
\begin{figure}[!h]
    \centering
    \begin{subfigure}[b]{0.42\textwidth}
         \centering
         \begin{tikzpicture}
\small
    \begin{axis}[
        height=1.4in,
        width=\linewidth,
        title={Github},
        xlabel={\% Edges ($\cdot 100$)},
        ylabel={Accuracy},
        xmin=0, xmax=1,
        xtick=\myxticks,
        xticklabel style={font=\scriptsize},
        yticklabel style={font=\scriptsize},
        ymajorgrids=true,
        xmajorgrids=true,
        grid style=dashed,
        enlargelimits=0.1,
        legend pos=south west,
        label style={font=\footnotesize},
    ]
     
    \addplot[
        color=blue,mark=square,style=solid,line width=1pt
        ]
        coordinates {
(0.1, 0.8568859416445623)
(0.2, 0.8572944297082228)
(0.3, 0.8595490716180372)
(0.4, 0.8604668435013263)
(0.5, 0.8608753315649867)
(0.6, 0.8623342175066313)
(0.7, 0.862710875331565)
(0.8, 0.863129973474801)
(0.9, 0.8649867374005306)
(1.0, 0.8692307692307693)
        };
        
        \addplot[
        color=purple,mark=pentagon,style=solid,line width=1pt
        ]
        coordinates {
(0.06522423642661149, 0.8587533156498673)
(0.3121022965159531, 0.860946949602122)
(0.4406909270838019, 0.8620291777188329)
(0.5221520191831919, 0.8631511936339522)
(0.5790562727722549, 0.8639363395225464)
(0.6213932035307592, 0.8645278514588859)
(0.6543028965097248, 0.8652122015915119)
(0.6809738999249143, 0.8655384615384616)
(0.702819693913212, 0.8663342175066313)
(0.7212624090407366, 0.8665278514588859)
(0.7370892343678094, 0.866129973474801)
(0.7370892343678094, 0.8653236074270557)
(0.7507136604118296, 0.866132095490716)
(0.7626616332702428, 0.8668859416445623)
(0.7731822853049969, 0.8671604774535809)
(0.7824953374186427, 0.8676931034482759)
(0.7908360812863534, 0.8678257294429709)
(0.8049968339429002, 0.8679151193633953)
(0.8193271350124394, 0.8679572413793103)
(0.8310155949938236, 0.8680120954907162)
(0.8406227617014356, 0.8680151193633953)
(0.8515572670783466, 0.8684749236743992)
(0.8571122265538028, 0.8687735238238737)
(0.8668741725850254, 0.868941674138226)
        };
        
        \addplot[
        color=black,mark=diamond,style=solid,line width=1pt
        ]
        coordinates {
(0.06522423642661149, 0.8571007957559682)
(0.3121022965159531, 0.862846949602122)
(0.4406909270838019, 0.8643291777188329)
(0.5221520191831919, 0.8651511936339522)
(0.5790562727722549, 0.8659363395225464)
(0.6213932035307592, 0.8657278514588859)
(0.6543028965097248, 0.8664122015915119)
(0.6809738999249143, 0.8672384615384616)
(0.7028196939132121, 0.8671342175066313)
(0.7212624090407366, 0.86705278514588859)
(0.7370892343678094, 0.8670129973474801)
(0.7507136604118296, 0.866932095490716)
(0.7626616332702428, 0.8671859416445623)
(0.7824953374186427, 0.8676931034482759)
(0.8049968339429002, 0.8679151193633953)
(0.8310155949938236, 0.8683120954907162)
(0.8571122265538028, 0.8688735238238737)
(0.8668741725850254, 0.868941674138226)
        };
    \end{axis}
    
    \begin{axis}[
        height=1.4in,
        width=\linewidth,
        axis y line=right,
        xmin=0, xmax=1,
        ymin=0.5, ymax=1,
        xticklabel style={font=\scriptsize},
        yticklabel style={font=\scriptsize},
        ylabel={\timelabel},
        xtick=\myxticks,
        enlargelimits=0.1,
        label style={font=\footnotesize},
    ]
        \addplot[color=teal, style=densely dotted, line width=2pt]
          coordinates{
(0.1, 0.7522020594874795)
(0.2, 0.7721805566177885)
(0.3, 0.8040854565093087)
(0.4, 0.8232223847521216)
(0.5, 0.852340634596584)
(0.6, 0.8702750458955674)
(0.7, 0.909085750906387)
(0.8, 0.9228992443490338)
(0.9, 0.9470032720028777)
(1.0, 1.0)
        };

        

    \end{axis}
    \end{tikzpicture}
     \end{subfigure}\qquad\qquad
     \begin{subfigure}[b]{\textwidth}
         \centering
         \begin{tikzpicture}
\small
    \begin{axis}[
        axis lines=none,
        height=0.8in,
        width=0.8in,
        title={},
        xmin=0, xmax=1,
        ymin=0, ymax=1,
        ytick={},
        xtick={},
        legend columns=6,
        legend style={font=\footnotesize, anchor=south east,at={(1,0)},nodes={scale=0.8, transform shape}},
    ]
     \addplot+[forget plot]
       coordinates{
   (0, 2.388854806)
       };
    \addlegendimage{color=blue,mark=square,style=solid,line width=1pt}
    \addlegendentry{Random}
    \addlegendimage{color=purple,mark=pentagon,style=solid,line width=1pt}
    \addlegendentry{LSP-P}
    \addlegendimage{color=black,mark=diamond,style=solid,line width=1pt}
    \addlegendentry{LSP-T}
    \addlegendimage{color=teal, style=densely dotted, line width=2pt}
    \addlegendentry{Relative Testing Time}
    \end{axis}
    \end{tikzpicture}
     \end{subfigure}
    \caption{Additional results for node classification problems. We report the results in terms of the metric identified with the dataset (in solid lines), and the running time per iteration (in densely dotted lines) under the specified pruning configurations.}
    \label{fig:add_node_classification_results}
\end{figure}

Additional results for node classification benchmarks are presented in Figure \ref{fig:add_node_classification_results}.

\subsection{Graph Regression}
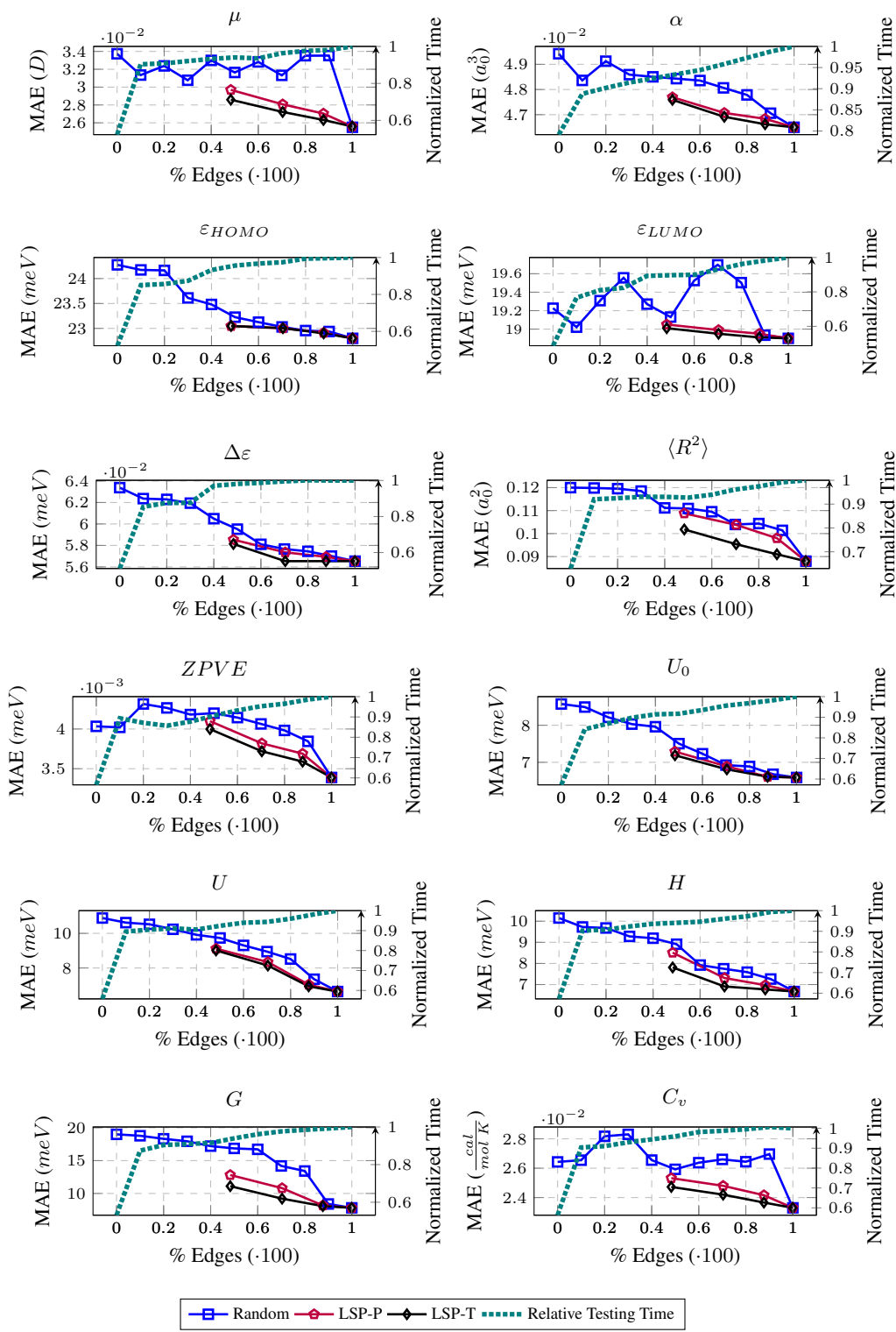
\begin{figure}[!ht]
\footnotesize
    \centering
    \begin{subfigure}[b]{0.42\textwidth}
         \centering
         \begin{tikzpicture}
\small
    \begin{axis}[
        height=\qmheight,
        width=\linewidth,
        title={$\mu$},
        xlabel={\% Edges ($\cdot 100$)},
        ylabel={MAE ($D$)},
        xtick=\myxticks,
        ymajorgrids=true,
        xmajorgrids=true,
        grid style=dashed,
        enlargelimits=0.1,
        legend pos=south west,
        xticklabel style={font=\scriptsize},
        yticklabel style={font=\scriptsize},
        label style={font=\footnotesize},
    ]
     
    \addplot[
        color=blue,mark=square,style=solid,line width=1pt
        ]
        coordinates {
(0.0, 0.033735979021088505)
(0.1, 0.031349508989965726)
(0.2, 0.03237473956413305)
(0.3, 0.030768171834866225)
(0.4, 0.03299352678628616)
(0.5, 0.031641624874995056)
(0.6, 0.03282199324140636)
(0.7, 0.03131845086611739)
(0.8, 0.03351728260013704)
(0.9, 0.033531075667383364)
(1.0, 0.0255)
        };
        
        \addplot[
        color=purple,mark=pentagon,style=solid,line width=1pt
        ]
        coordinates {
(0.48318212868875066, 0.029693041048877256)
(0.7035288004586339, 0.02807081115592106)
(0.8768405079558135, 0.0270357962469907)
(0.9994301909075598, 0.025589568202613155)
        };
        
        \addplot[
        color=black,mark=diamond,style=solid,line width=1pt
        ]
        coordinates {
(0.48318212868875066, 0.0285888435474105)
(0.7035288004586339, 0.027216341177427744)
(0.8768405079558135, 0.026321315510747508)
(0.9994301909075598, 0.025589568202613155)
        };
    \end{axis}
    
    \begin{axis}[
        height=\qmheight,
        width=\linewidth,
        axis y line=right,
        xticklabel style={font=\scriptsize},
        yticklabel style={font=\scriptsize},
        ylabel={\timelabel},
        xtick=\myxticks,
        label style={font=\footnotesize},
    ]
        \addplot[color=teal, style=densely dotted, line width=2pt]
          coordinates{

(0.0, 0.5229324280744425)
(0.1, 0.9033374756138199)
(0.2, 0.908055830409958)
(0.3, 0.9207097496247466)
(0.4, 0.9326186614300677)
(0.5, 0.9404050326687592)
(0.6, 0.93553133752168)
(0.7, 0.9629465579994054)
(0.8, 0.9759265067151759)
(0.9, 0.9803781473953123)
(1.0, 1.0)
        };

        

    \end{axis}
    \end{tikzpicture}
     \end{subfigure}\hspace{20pt}
    \begin{subfigure}[b]{0.42\textwidth}
         \centering
         \begin{tikzpicture}
\small
    \begin{axis}[
        height=\qmheight,
        width=\linewidth,
        title={$\alpha$},
        xlabel={\% Edges ($\cdot 100$)},
        ylabel={MAE ($a_0^3$)},
        xtick=\myxticks,
        xticklabel style={font=\scriptsize},
        yticklabel style={font=\scriptsize},
        ymajorgrids=true,
        xmajorgrids=true,
        grid style=dashed,
        legend pos=south west,
        label style={font=\footnotesize},
    ]
     
    \addplot[
        color=blue,mark=square,style=solid,line width=1pt
        ]
        coordinates {
(0.0, 0.04941887086980786)
(0.1, 0.04836113156799334)
(0.2, 0.04912372508988532)
(0.3, 0.04859003933287144)
(0.4, 0.048502494517304554)
(0.5, 0.04843069450029299)
(0.6, 0.04835587180465846)
(0.7, 0.0480647125201423)
(0.8, 0.04778478964368619)
(0.9, 0.04705872848571686)
(1.0, 0.0465)
        };
        
        \addplot[
        color=purple,mark=pentagon,style=solid,line width=1pt
        ]
        coordinates {
(0.48318212868875066, 0.047693041048877256)
(0.7035288004586339, 0.04707081115592106)
(0.8768405079558135, 0.0468357962469907)
(0.9994301909075598, 0.0465)
        };
        
        \addplot[
        color=black,mark=diamond,style=solid,line width=1pt
        ]
        coordinates {
(0.48318212868875066, 0.0475888435474105)
(0.7035288004586339, 0.046916341177427744)
(0.8768405079558135, 0.046621315510747508)
(0.9994301909075598, 0.0465)
        };
    \end{axis}
    
    \begin{axis}[
        height=\qmheight,
        width=\linewidth,
        axis y line=right,
        xticklabel style={font=\scriptsize},
        yticklabel style={font=\scriptsize},
        ylabel={\timelabel},
        xtick=\myxticks,
        label style={font=\footnotesize},
    ]
        \addplot[color=teal, style=densely dotted, line width=2pt]
          coordinates{
(0.0, 0.7913797984934976)
(0.1, 0.8878655742337983)
(0.2, 0.901866084835368)
(0.3, 0.9145762744644711)
(0.4, 0.9242231156143511)
(0.5, 0.9341590166798548)
(0.6, 0.9440394241660256)
(0.7, 0.9578259502967902)
(0.8, 0.9729441747565098)
(0.9, 0.9877495716705057)
(1.0, 1.0)
        };

        

    \end{axis}
    \end{tikzpicture}
     \end{subfigure}
     
     \hspace{-10pt}
     \begin{subfigure}[b]{0.42\textwidth}
         \centering
         \begin{tikzpicture}
\small
    \begin{axis}[
        height=\qmheight,
        width=\linewidth,
        title={$\varepsilon_{HOMO}$},
        xlabel={\% Edges ($\cdot 100$)},
        ylabel={MAE ($meV$)},
        xtick=\myxticks,
        xticklabel style={font=\scriptsize},
        yticklabel style={font=\scriptsize},
        ymajorgrids=true,
        xmajorgrids=true,
        grid style=dashed,
        legend pos=south west,
        label style={font=\footnotesize},
    ]
     
    \addplot[
        color=blue,mark=square,style=solid,line width=1pt
        ]
        coordinates {
(0.0, 24.275498491392757)
(0.1, 24.173422336790575)
(0.2, 24.164015128568418)
(0.3, 23.611847426416404)
(0.4, 23.48064005273118)
(0.5, 23.2253895841452)
(0.6, 23.124926740730223)
(0.7, 23.02956476884448)
(0.8, 22.956053054873157)
(0.9, 22.93754230384263)
(1.0, 22.8)
        };
        
        \addplot[
        color=purple,mark=pentagon,style=solid,line width=1pt
        ]
        coordinates {
(0.48318212868875066, 23.04819301949)
(0.7035288004586339, 23.00193018409)
(0.8768405079558135, 22.9118832613)
(0.9994301909075598, 22.8)
        };
        
        \addplot[
        color=black,mark=diamond,style=solid,line width=1pt
        ]
        coordinates {
(0.48318212868875066, 23.04819301949)
(0.7035288004586339, 23.00514616311)
(0.8768405079558135, 22.9001990214)
(0.9994301909075598, 22.8)
        };
    \end{axis}
    
    \begin{axis}[
        height=\qmheight,
        width=\linewidth,
        axis y line=right,
        xticklabel style={font=\scriptsize},
        yticklabel style={font=\scriptsize},
        ylabel={\timelabel},
        xtick=\myxticks,
        label style={font=\footnotesize},
    ]
        \addplot[color=teal, style=densely dotted, line width=2pt]
          coordinates{
(0.0, 0.5222965429727972)
(0.1, 0.8506134872782345)
(0.2, 0.8563102993372134)
(0.3, 0.8736668178289307)
(0.4, 0.9322311935533293)
(0.5, 0.9562229361369817)
(0.6, 0.9671524308461321)
(0.7, 0.9740023155568545)
(0.8, 0.9935440342602976)
(0.9, 0.9965525523564515)
(1.0, 1.0)
        };

        

    \end{axis}
    \end{tikzpicture}
     \end{subfigure}\hspace{20pt}
     \begin{subfigure}[b]{0.42\textwidth}
         \centering
         \begin{tikzpicture}
\small
    \begin{axis}[
        height=\qmheight,
        width=\linewidth,
        title={$\varepsilon_{LUMO}$},
        xlabel={\% Edges ($\cdot 100$)},
        ylabel={MAE ($meV$)},
        xtick=\myxticks,
        xticklabel style={font=\scriptsize},
        yticklabel style={font=\scriptsize},
        ymajorgrids=true,
        xmajorgrids=true,
        grid style=dashed,
        legend pos=south west,
        yticklabel style = {font=\tiny},
        label style={font=\footnotesize},
    ]
     
    \addplot[
        color=blue,mark=square,style=solid,line width=1pt
        ]
        coordinates {
(0.0, 19.226271754831647)
(0.1, 19.021299888288183)
(0.2, 19.3081934238261)
(0.3, 19.55557473828481)
(0.4, 19.270547668946705)
(0.5, 19.130625427855687)
(0.6, 19.526104403634)
(0.7, 19.696222668446396)
(0.8, 19.50341266461965)
(0.9, 18.934323063138446)
(1.0, 18.9)
        };
        
        \addplot[
        color=purple,mark=pentagon,style=solid,line width=1pt
        ]
        coordinates {
(0.48318212868875066, 19.05)
(0.7035288004586339, 18.99)
(0.8768405079558135, 18.95)
(0.9994301909075598, 18.9)
        };
        
        \addplot[
        color=black,mark=diamond,style=solid,line width=1pt
        ]
        coordinates {
(0.48318212868875066, 19.01)
(0.7035288004586339, 18.95)
(0.8768405079558135, 18.91)
(0.9994301909075598, 18.9)
        };
    \end{axis}
    
    \begin{axis}[
        height=\qmheight,
        width=\linewidth,
        axis y line=right,
        xticklabel style={font=\scriptsize},
        yticklabel style={font=\scriptsize},
        ylabel={\timelabel},
        xtick=\myxticks,
        label style={font=\footnotesize},
    ]
        \addplot[color=teal, style=densely dotted, line width=2pt]
          coordinates{
(0.0, 0.4873105837954644)
(0.1, 0.7662416017042086)
(0.2, 0.8103188278498276)
(0.3, 0.8239252211192559)
(0.4, 0.8932681961472287)
(0.5, 0.8970563917483234)
(0.6, 0.8995658154542358)
(0.7, 0.9300841192911113)
(0.8, 0.9617910511945502)
(0.9, 0.9820315903987714)
(1.0, 1.0)
        };

        

    \end{axis}
    \end{tikzpicture}
     \end{subfigure}
     
     \hspace{-0pt}
    \begin{subfigure}[b]{0.42\textwidth}
         \centering
         \begin{tikzpicture}
\small
    \begin{axis}[
        height=\qmheight,
        width=\linewidth,
        title={$\Delta\varepsilon$},
        xlabel={\% Edges ($\cdot 100$)},
        ylabel={MAE ($meV$)},
        xtick=\myxticks,
        xticklabel style={font=\scriptsize},
        yticklabel style={font=\scriptsize},
        ymajorgrids=true,
        xmajorgrids=true,
        grid style=dashed,
        legend pos=south west,
        label style={font=\footnotesize},
    ]
     
    \addplot[
        color=blue,mark=square,style=solid,line width=1pt
        ]
        coordinates {
(0,  0.06338372257351875 )
(0.1,  0.06236105039715767 )
(0.2,  0.06228857898712158 )
(0.3,  0.06193902009725571 )
(0.4,  0.060491565614938736 )
(0.5,  0.059515005856752396 )
(0.6,  0.058104981780052185 )
(0.7,  0.0576662773668766 )
(0.8,  0.057447512394189835 )
(0.9,  0.05700638653635979 )
(1.0,  0.05653313174843788 )
        };
        
        \addplot[
        color=purple,mark=pentagon,style=solid,line width=1pt
        ]
        coordinates {
(0.48318212868875066, 0.05853313174843788)
(0.7035288004586339, 0.057353313174843788)
(0.8768405079558135, 0.05693313174843788)
(0.9994301909075598, 0.05653313174843788)
        };
        
        \addplot[
        color=black,mark=diamond,style=solid,line width=1pt
        ]
        coordinates {
(0.48318212868875066, 0.05813313174843788)
(0.7035288004586339, 0.05653313174843788)
(0.8768405079558135, 0.05653313174843788)
(0.9994301909075598, 0.05653313174843788)
        };
    \end{axis}
    
    \begin{axis}[
        height=\qmheight,
        width=\linewidth,
        axis y line=right,
        xticklabel style={font=\scriptsize},
        yticklabel style={font=\scriptsize},
        ylabel={\timelabel},
        xtick=\myxticks,
        label style={font=\footnotesize},
    ]
        \addplot[color=teal, style=densely dotted, line width=2pt]
          coordinates{
(0.0, 0.5114017162210397)
(0.1, 0.853941579709251)
(0.2, 0.8755679343063154)
(0.3, 0.8756003297268774)
(0.4, 0.970520823922672)
(0.5, 0.9807772990084355)
(0.6, 0.9871688907890499)
(0.7, 0.9950410252382114)
(0.8, 1.001016234806302)
(0.9, 0.9999192220053076)
(1.0, 1.0)
        };

        

    \end{axis}
    \end{tikzpicture}
     \end{subfigure}\hspace{20pt}
     \begin{subfigure}[b]{0.42\textwidth}
         \centering
         \begin{tikzpicture}
\small
    \begin{axis}[
        height=\qmheight,
        width=\linewidth,
        title={$\langle R^2 \rangle$},
        xlabel={\% Edges ($\cdot 100$)},
        ylabel={MAE ($a_0^2$)},
        xtick=\myxticks,
        xticklabel style={font=\scriptsize},
        yticklabel style={font=\scriptsize},
        ymajorgrids=true,
        xmajorgrids=true,
        grid style=dashed,
        legend pos=south west,
        scaled ticks=false, tick label style={/pgf/number format/fixed},
        label style={font=\footnotesize},
    ]
     
    \addplot[
        color=blue,mark=square,style=solid,line width=1pt
        ]
        coordinates {
(0.0, 0.119991767345253892)
(0.1, 0.119834151319211305)
(0.2, 0.11953852672632555)
(0.3, 0.118502574046133572)
(0.4, 0.11117214077402625)
(0.5, 0.11091209044652244)
(0.6, 0.10950783170398339)
(0.7, 0.10395677048413035)
(0.8, 0.10435918566519193)
(0.9, 0.10140476228546371)
(1.0, 0.088)
        };
        
        \addplot[
        color=purple,mark=pentagon,style=solid,line width=1pt
        ]
        coordinates {
(0.48318212868875066, 0.1088)
(0.7035288004586339, 0.1038)
(0.8768405079558135, 0.098)
(0.9994301909075598, 0.088)
        };
        
        \addplot[
        color=black,mark=diamond,style=solid,line width=1pt
        ]
        coordinates {
(0.48318212868875066, 0.1018)
(0.7035288004586339, 0.09538)
(0.8768405079558135, 0.091)
(0.9994301909075598, 0.088)
        };
    \end{axis}
    
    \begin{axis}[
        height=\qmheight,
        width=\linewidth,
        axis y line=right,
        xticklabel style={font=\scriptsize},
        yticklabel style={font=\scriptsize},
        ylabel={\timelabel},
        xtick=\myxticks,
        label style={font=\footnotesize},
    ]
        \addplot[color=teal, style=densely dotted, line width=2pt]
          coordinates{
(0.0, 0.629801656685559)
(0.1, 0.92051001738778)
(0.2, 0.9247373421423465)
(0.3, 0.9304687284438243)
(0.4, 0.9305439026829008)
(0.5, 0.9272327938786817)
(0.6, 0.9395066897847822)
(0.7, 0.9612973875353502)
(0.8, 0.9743923789532576)
(0.9, 0.9916368269386812)
(1.0, 1.0)
        };

        

    \end{axis}
    \end{tikzpicture}
     \end{subfigure}
     
     \begin{subfigure}[b]{0.42\textwidth}
         \centering
         \begin{tikzpicture}
\small
    \begin{axis}[
        height=\qmheight,
        width=\linewidth,
        title={$ZPVE$},
        xlabel={\% Edges ($\cdot 100$)},
        ylabel={MAE ($meV$)},
        xtick=\myxticks,
        xticklabel style={font=\scriptsize},
        yticklabel style={font=\scriptsize},
        ymajorgrids=true,
        xmajorgrids=true,
        grid style=dashed,
        legend pos=south west,
        label style={font=\footnotesize},
    ]
     
    \addplot[
        color=blue,mark=square,style=solid,line width=1pt
        ]
        coordinates {
(0.0, 0.0040335374492941065)
(0.1, 0.0040205432269364155)
(0.2, 0.004315326687311438)
(0.3, 0.004265699046813754)
(0.4, 0.0041819382241202584)
(0.5, 0.004199046842530533)
(0.6, 0.0041438555821704408)
(0.7, 0.00406394089992978)
(0.8, 0.003982184067677927)
(0.9, 0.003844640836053498)
(1.0, 0.0033893206242182634)
        };
        
        \addplot[
        color=purple,mark=pentagon,style=solid,line width=1pt
        ]
        coordinates {
(0.48318212868875066, 0.004099046842530533)
(0.7035288004586339, 0.0038193206242182634)
(0.8768405079558135, 0.0036893206242182634)
(0.9994301909075598, 0.0033893206242182634)
        };
        
        \addplot[
        color=black,mark=diamond,style=solid,line width=1pt
        ]
        coordinates {
(0.48318212868875066, 0.003999046842530533)
(0.7035288004586339, 0.0037193206242182634)
(0.8768405079558135, 0.0035893206242182634)
(0.9994301909075598, 0.0033893206242182634)
        };
    \end{axis}
    
    \begin{axis}[
        height=\qmheight,
        width=\linewidth,
        axis y line=right,
        xticklabel style={font=\scriptsize},
        yticklabel style={font=\scriptsize},
        ylabel={\timelabel},
        xtick=\myxticks,
        label style={font=\footnotesize},
    ]
        \addplot[color=teal, style=densely dotted, line width=2pt]
          coordinates{
(0.0, 0.56582904369246)
(0.1, 0.8934217732969282)
(0.2, 0.8712202339819798)
(0.3, 0.8567264064450499)
(0.4, 0.8772814832258915)
(0.5, 0.9059628033777931)
(0.6, 0.9322626561321895)
(0.7, 0.9527117282015789)
(0.8, 0.9650456328359261)
(0.9, 0.9856012351146148)
(1.0, 1.0)
        };

        

    \end{axis}
    \end{tikzpicture}
     \end{subfigure}\qquad\hspace{20pt}
    \begin{subfigure}[b]{0.42\textwidth}
         \centering
         \begin{tikzpicture}
\small
    \begin{axis}[
        height=\qmheight,
        width=\linewidth,
        title={$U_0$},
        xlabel={\% Edges ($\cdot 100$)},
        ylabel={MAE ($meV$)},
        xtick=\myxticks,
        xticklabel style={font=\scriptsize},
        yticklabel style={font=\scriptsize},
        ymajorgrids=true,
        xmajorgrids=true,
        grid style=dashed,
        legend pos=south west,
        label style={font=\footnotesize},
    ]
     
    \addplot[
        color=blue,mark=square,style=solid,line width=1pt
        ]
        coordinates {
(0.0, 8.574519618994389)
(0.1, 8.491580208684238)
(0.2, 8.213679770110095)
(0.3, 8.030991347685518)
(0.4, 7.960026312347367)
(0.5, 7.50359736382743)
(0.6, 7.2311609973012265)
(0.7, 6.924577478692837)
(0.8, 6.8877676731466595)
(0.9, 6.678316484791061)
(1.0, 6.59)
        };
        
        \addplot[
        color=purple,mark=pentagon,style=solid,line width=1pt
        ]
        coordinates {
(0.48318212868875066, 7.29)
(0.7035288004586339, 6.89)
(0.8768405079558135, 6.61)
(0.9994301909075598, 6.59)
        };
        
        \addplot[
        color=black,mark=diamond,style=solid,line width=1pt
        ]
        coordinates {
(0.48318212868875066, 7.19)
(0.7035288004586339, 6.81)
(0.8768405079558135, 6.61)
(0.9994301909075598, 6.59)
        };
    \end{axis}
    
    \begin{axis}[
        height=\qmheight,
        width=\linewidth,
        axis y line=right,
        xticklabel style={font=\scriptsize},
        yticklabel style={font=\scriptsize},
        ylabel={\timelabel},
        xtick=\myxticks,
        label style={font=\footnotesize},
    ]
        \addplot[color=teal, style=densely dotted, line width=2pt]
          coordinates{
(0.0, 0.5719097820723692)
(0.1, 0.8421039859677893)
(0.2, 0.8696470563298008)
(0.3, 0.8977457498564185)
(0.4, 0.9140507732921317)
(0.5, 0.9170698104943455)
(0.6, 0.9351506535334324)
(0.7, 0.9562101898743032)
(0.8, 0.9684938294051696)
(0.9, 0.9823539557430317)
(1.0, 1.0)
        };

        

    \end{axis}
    \end{tikzpicture}
     \end{subfigure}
     
     \begin{subfigure}[b]{0.42\textwidth}
         \centering
         \begin{tikzpicture}
\small
    \begin{axis}[
        height=\qmheight,
        width=\linewidth,
        title={$U$},
        xlabel={\% Edges ($\cdot 100$)},
        ylabel={MAE ($meV$)},
        xtick=\myxticks,
        xticklabel style={font=\scriptsize},
        yticklabel style={font=\scriptsize},
        ymajorgrids=true,
        xmajorgrids=true,
        grid style=dashed,
        legend pos=south west,
        label style={font=\footnotesize},
    ]
     
    \addplot[
        color=blue,mark=square,style=solid,line width=1pt
        ]
        coordinates {
(0.0, 10.885976440975945)
(0.1, 10.619130020710439)
(0.2, 10.523634162372597)
(0.3, 10.234608862451991)
(0.4, 9.914679413068278)
(0.5, 9.7332120264956785)
(0.6, 9.304649894056442)
(0.7, 8.94479810282132)
(0.8, 8.518337707308506)
(0.9, 7.350959008406375)
(1.0, 6.64)
        };
        
        \addplot[
        color=purple,mark=pentagon,style=solid,line width=1pt
        ]
        coordinates {
(0.48318212868875066, 9.1132120264956785)
(0.7035288004586339, 8.34)
(0.8768405079558135, 7.04)
(0.9994301909075598, 6.64)
        };
        
        \addplot[
        color=black,mark=diamond,style=solid,line width=1pt
        ]
        coordinates {
(0.48318212868875066, 9.0132120264956785)
(0.7035288004586339, 8.14)
(0.8768405079558135, 6.94)
(0.9994301909075598, 6.64)
        };
    \end{axis}
    
    \begin{axis}[
        height=\qmheight,
        width=\linewidth,
        axis y line=right,
        xticklabel style={font=\scriptsize},
        yticklabel style={font=\scriptsize},
        ylabel={\timelabel},
        xtick=\myxticks,
        label style={font=\footnotesize},
    ]
        \addplot[color=teal, style=densely dotted, line width=2pt]
          coordinates{
(0.0, 0.5576742971626373)
(0.1, 0.8952999306978472)
(0.2, 0.904638598336335)
(0.3, 0.9109840596853935)
(0.4, 0.9050423628701748)
(0.5, 0.9238829953489853)
(0.6, 0.9394734282917032)
(0.7, 0.9436525700315112)
(0.8, 0.95943424788821)
(0.9, 0.9815796423626209)
(1.0, 1.0)
        };

        

    \end{axis}
    \end{tikzpicture}
     \end{subfigure}\qquad\hspace{10pt}
     \begin{subfigure}[b]{0.42\textwidth}
         \centering
         \begin{tikzpicture}
\small
    \begin{axis}[
        height=\qmheight,
        width=\linewidth,
        title={$H$},
        xlabel={\% Edges ($\cdot 100$)},
        ylabel={MAE ($meV$)},
        xtick=\myxticks,
        xticklabel style={font=\scriptsize},
        yticklabel style={font=\scriptsize},
        ymajorgrids=true,
        xmajorgrids=true,
        grid style=dashed,
        legend pos=south west,
        label style={font=\footnotesize},
    ]
     
    \addplot[
        color=blue,mark=square,style=solid,line width=1pt
        ]
        coordinates {
(0.0, 10.150545198081853)
(0.1, 9.7253330286268955)
(0.2, 9.677968428881589)
(0.3, 9.2799252630508966)
(0.4, 9.193946307480531)
(0.5, 8.910553120210627)
(0.6, 7.9225641121911448)
(0.7, 7.74865660395727)
(0.8, 7.5872661211061345)
(0.9, 7.261595553385785)
(1.0, 6.67)
        };
        
        \addplot[
        color=purple,mark=pentagon,style=solid,line width=1pt
        ]
        coordinates {
(0.48318212868875066, 8.510553120210627)
(0.7035288004586339, 7.31)
(0.8768405079558135, 6.97)
(0.9994301909075598, 6.67)
        };
        
        \addplot[
        color=black,mark=diamond,style=solid,line width=1pt
        ]
        coordinates {
(0.48318212868875066, 7.810553120210627)
(0.7035288004586339, 6.91)
(0.8768405079558135, 6.77)
(0.9994301909075598, 6.67)
        };
    \end{axis}
    
    \begin{axis}[
        height=\qmheight,
        width=\linewidth,
        axis y line=right,
        xticklabel style={font=\scriptsize},
        yticklabel style={font=\scriptsize},
        ylabel={\timelabel},
        xtick=\myxticks,
        label style={font=\footnotesize},
    ]
        \addplot[color=teal, style=densely dotted, line width=2pt]
          coordinates{
(0.0, 0.5725213553321962)
(0.1, 0.9033413143227832)
(0.2, 0.9076788154085256)
(0.3, 0.9234103045359751)
(0.4, 0.9361929423508141)
(0.5, 0.9409575974966532)
(0.6, 0.9463002297734058)
(0.7, 0.9604588811389946)
(0.8, 0.9727186624155589)
(0.9, 0.9936249744055392)
(1.0, 1.0)
        };

        

    \end{axis}
    \end{tikzpicture}
     \end{subfigure}
     
    \begin{subfigure}[b]{0.42\textwidth}
         \centering
         \begin{tikzpicture}
\small
    \begin{axis}[
        height=\qmheight,
        width=\linewidth,
        title={$G$},
        xlabel={\% Edges ($\cdot 100$)},
        ylabel={MAE ($meV$)},
        xtick=\myxticks,
        xticklabel style={font=\scriptsize},
        yticklabel style={font=\scriptsize},
        ymajorgrids=true,
        xmajorgrids=true,
        grid style=dashed,
        legend pos=south west,
        label style={font=\footnotesize},
    ]
     
    \addplot[
        color=blue,mark=square,style=solid,line width=1pt
        ]
        coordinates {
(0.0, 18.99655228569506)
(0.1, 18.77930957875433)
(0.2, 18.317121194565063)
(0.3, 17.942050181143877)
(0.4, 17.205188198510583)
(0.5, 16.856327502904758)
(0.6, 16.715962010508907)
(0.7, 14.188370190903404)
(0.8, 13.41528380908421)
(0.9, 8.399110690962156)
(1.0, 7.809999999999999)
        };
        
        \addplot[
        color=purple,mark=pentagon,style=solid,line width=1pt
        ]
        coordinates {
(0.48318212868875066, 12.809999999999999)
(0.7035288004586339, 10.809999999999999)
(0.8768405079558135, 8.179999999999999)
(0.9994301909075598, 7.809999999999999)
        };
        
        \addplot[
        color=black,mark=diamond,style=solid,line width=1pt
        ]
        coordinates {
(0.48318212868875066, 11.109999999999999)
(0.7035288004586339, 9.209999999999999)
(0.8768405079558135, 8.079999999999999)
(0.9994301909075598, 7.809999999999999)
        };
    \end{axis}
    
    \begin{axis}[
        height=\qmheight,
        width=\linewidth,
        axis y line=right,
        xticklabel style={font=\scriptsize},
        yticklabel style={font=\scriptsize},
        ylabel={\timelabel},
        xtick=\myxticks,
        label style={font=\footnotesize},
    ]
        \addplot[color=teal, style=densely dotted, line width=2pt]
          coordinates{
(0.0, 0.5309374433268415)
(0.1, 0.8740831285534341)
(0.2, 0.9038800981616036)
(0.3, 0.9093686232782956)
(0.4, 0.9150149594253163)
(0.5, 0.939677741140712)
(0.6, 0.9613473066011141)
(0.7, 0.976500498360379)
(0.8, 0.9847076725372563)
(0.9, 0.9920162098368355)
(1.0, 1.0)
        };

        

    \end{axis}
    \end{tikzpicture}
     \end{subfigure}\qquad
     \begin{subfigure}[b]{0.42\textwidth}
         \centering
         \begin{tikzpicture}
\small
    \begin{axis}[
        height=\qmheight,
        width=\linewidth,
        title={$C_v$},
        xlabel={\% Edges ($\cdot 100$)},
        ylabel={MAE ($\frac{cal}{mol\ K}$)},
        xtick=\myxticks,
        xticklabel style={font=\scriptsize},
        yticklabel style={font=\scriptsize},
        ymajorgrids=true,
        xmajorgrids=true,
        grid style=dashed,
        legend pos=south west,
        label style={font=\footnotesize},
    ]
     
    \addplot[
        color=blue,mark=square,style=solid,line width=1pt
        ]
        coordinates {
(0.0, 0.026428666006987074)
(0.1, 0.026546667436393088)
(0.2, 0.028173385614594978)
(0.3, 0.028306079425543553)
(0.4, 0.026552543828553064)
(0.5, 0.025923867947419268)
(0.6, 0.026375503853593225)
(0.7, 0.026600317187419665)
(0.8, 0.026440996062067613)
(0.9, 0.026961618626417192)
(1.0, 0.0233)
        };
        
        \addplot[
        color=purple,mark=pentagon,style=solid,line width=1pt
        ]
        coordinates {
(0.48318212868875066, 0.025323867947419268)
(0.7035288004586339, 0.024800317187419665)
(0.8768405079558135, 0.024161618626417192)
(0.9994301909075598, 0.0233)
        };
        
        \addplot[
        color=black,mark=diamond,style=solid,line width=1pt
        ]
        coordinates {
(0.48318212868875066, 0.024723867947419268)
(0.7035288004586339, 0.024200317187419665)
(0.8768405079558135, 0.023661618626417192)
(0.9994301909075598, 0.0233)
        };
    \end{axis}
    
    \begin{axis}[
        height=\qmheight,
        width=\linewidth,
        axis y line=right,
        xticklabel style={font=\scriptsize},
        yticklabel style={font=\scriptsize},
        ylabel={\timelabel},
        xtick=\myxticks,
        label style={font=\footnotesize},
    ]
        \addplot[color=teal, style=densely dotted, line width=2pt]
          coordinates{
(0.0, 0.5630415751948992)
(0.1, 0.9037502061200253)
(0.2, 0.9105137719622534)
(0.3, 0.9295714346819832)
(0.4, 0.9445666579914009)
(0.5, 0.9595532047936672)
(0.6, 0.9813899095436194)
(0.7, 0.9874753407729037)
(0.8, 0.9952490537789477)
(0.9, 1.0067510474629153)
(1.0, 1.0)
        };

        

    \end{axis}
    \end{tikzpicture}
     \end{subfigure}
     
     \begin{subfigure}[b]{\textwidth}
         \centering
         \begin{tikzpicture}
\small
    \begin{axis}[
        axis lines=none,
        height=0.8in,
        width=0.8in,
        title={},
        xmin=0, xmax=1,
        ymin=0, ymax=1,
        ytick={},
        xtick={},
        legend columns=6,
        legend style={font=\footnotesize, anchor=south east,at={(1,0)},nodes={scale=0.8, transform shape}},
    ]
     \addplot+[forget plot]
       coordinates{
   (0, 2.388854806)
       };
    \addlegendimage{color=blue,mark=square,style=solid,line width=1pt}
    \addlegendentry{Random}
    \addlegendimage{color=purple,mark=pentagon,style=solid,line width=1pt}
    \addlegendentry{LSP-P}
    \addlegendimage{color=black,mark=diamond,style=solid,line width=1pt}
    \addlegendentry{LSP-T}
    \addlegendimage{color=teal, style=densely dotted, line width=2pt}
    \addlegendentry{Relative Testing Time}
    \end{axis}
    \end{tikzpicture}
     \end{subfigure}
    \caption{MAE and normalized running times for additional QM9 targets. Errors are presented as solid lines (Lower is better). Normalized running times are presented as dotted lines.}
    \label{fig:additionl_graph_regression_results}
\end{figure}

Prediction results for all QM9 targets are presented in Figure \ref{fig:additionl_graph_regression_results}.


\end{document}